\ifdefined\pdfoutput
  \pdfoutput=1
\fi
\documentclass[11pt]{article}
\usepackage{times}
\usepackage[margin=1in]{geometry}
\usepackage[authoryear,round]{natbib}
\setcitestyle{authoryear,round,citesep={;},aysep={,},yysep={;}}

\usepackage[utf8]{inputenc}
\usepackage[T1]{fontenc}
\usepackage{hyperref}
\hypersetup{hidelinks}
\usepackage{url}
\usepackage{booktabs}
\usepackage{multirow}
\usepackage{amsfonts}
\usepackage{amsmath}
\usepackage{amssymb}
\usepackage{nicefrac}
\usepackage{microtype}
\usepackage{xcolor}
\usepackage{graphicx}
\usepackage{placeins}

\graphicspath{{figures/}}

\newcommand{\revcolor}{black}
\DeclareRobustCommand{\rev}[1]{\leavevmode{\color{\revcolor}#1}}
\newenvironment{revblock}{\par\color{\revcolor}}{\par}

\title{Mode Dynamics: A Framework for Neural Network Learning}
\author{Hossein Honarvar\\
\small Independent Researcher\\
\small\texttt{hoseinhonarvar@gmail.com}}
\date{}

\begin{document}
\maketitle

\begin{abstract}
We introduce Mode Dynamics, a framework that connects coordinated parameter motion during neural network learning to changes in model outputs.
Starting from a recorded parameter trajectory, we decompose its updates into collective modes and map each mode to its state-dependent output effect.
The framework organizes this link through three mechanisms.
\emph{Decomposition} identifies collective modes in parameter motion and their time-varying amplitudes.
\emph{Reconstruction} tests whether the effects induced by those modes recover local and accumulated output change.
\emph{Concentration} quantifies how behavioral contribution is distributed across modes.
Controlled scalar regression and next-token experiments directly test these mechanisms.
They show that small sets of leading modes reconstruct behavior, induced effects depend on the training path, and modal complexity varies with task and data.
\rev{In a controlled induction task, suppressing the leading modes during training impairs held-out accuracy, whereas matched controls do not.}
We then evaluate the framework on held-out image and language-model outputs using CIFAR-100 autoencoding and a public Pythia-70M pretraining trajectory.
Across these settings, the results connect compact modal subspaces of recorded parameter motion to evolving behavior.
\end{abstract}

\section{Introduction}

Understanding how neural networks learn remains difficult because optimization unfolds in a high-dimensional parameter space, while the object of interest is the learned function.
A central question is therefore not only whether gradient descent lowers loss, but which collective degrees of freedom drive changes in predictions over time.

Existing work provides two complementary perspectives.
Kernel and linearization analyses describe regimes in which function-space dynamics remain close to initialization \citep{jacot2018ntk,lee2019wide,chizat2019lazy}, whereas studies of feature learning emphasize that the parameter-to-function map can evolve substantially \citep{woodworth2020kernelrich,yang2021feature,mei2019meanfield,seleznova2022ntkbeyond,loo2022evolution}.
Separately, gradients, optimization trajectories, and adaptation updates often occupy low-dimensional subspaces \citep{gurari2018tiny,li2018intrinsic,li2021dldr,aghajanyan2021intrinsic,hu2021lora}.
These observations do not by themselves identify which trajectory directions carry behavioral change or whether their output effects remain fixed throughout training.

Mode Dynamics addresses this gap by treating parameter updates as discrete velocities.
\rev{Collective directions extracted from the recorded update trajectory serve as modes, along which many parameters across neurons and tensors move in fixed proportion, and their time-varying coefficients are amplitudes.}
Each mode maps to an induced output effect that can change as the network moves through parameter space.
Three linked mechanisms organize the description: \emph{decomposition} into a shared mode basis with checkpoint-dependent amplitudes, \emph{reconstruction} of one-step and accumulated output change from the realized modal contributions, and \emph{concentration} of behavioral energy across modes.
Reconstruction also tests \emph{path dependence} by comparing induced effects evaluated along the recorded trajectory with an initialization-frozen baseline.
\rev{Our main insight is that the output change produced by learning is organized in collective modes of parameter velocity: in the tested settings, a few modes carry most of it, and they need not be those with the most parameter motion.}

\paragraph{Contributions.}
Our main contributions are as follows.
\begin{itemize}
    \item We introduce Mode Dynamics, a trajectory-level framework that maps collective parameter motion to state-dependent output effects and organizes this link through three mechanisms: decomposition, reconstruction, and concentration.
    \item \rev{We show that mode-induced output effects are path dependent rather than fixed at initialization, as expected beyond the lazy regime \citep{chizat2019lazy,mei2018meanfield}, and that trajectory modes outperform dimension-matched random subspaces.}
    \item We evaluate concentration separately from reconstruction fidelity, since energy concentration and output fidelity need not coincide.
    \item \rev{We show in a controlled induction task that suppressing the leading modes during training impairs held-out accuracy, whereas energy- and tensor-matched controls do not.}
    \item We isolate these mechanisms in controlled regression and next-token tasks, then test transfer to CIFAR-100 autoencoding and a public Pythia-70M pretraining trajectory \citep{biderman2023pythia}, ranking modes on the Pile validation split and scoring without reranking on the untouched Pile test split \citep{gao2020pile}.
\end{itemize}

\section{Related Work}

\paragraph{Dynamical-systems perspectives.}
\rev{In physics, a collective coordinate tracks a pattern in which many degrees of freedom move together, and normal-mode analysis and dynamic mode decomposition describe high-dimensional motion through such patterns and their time-dependent amplitudes \citep{mcgaughey2013phonons,schmid2010dmd}.}
\rev{Koopman methods seek coordinates in which the dynamics become linear \citep{lusch2018koopman} and have been applied to neural-network training \citep{dogra2020koopman,manojlovic2021koopman}, while Correlation Mode Decomposition groups parameters whose values change in correlated ways \citep{brokman2024correlated}.}
\rev{Mode Dynamics instead finds collective coordinates directly in the recorded parameter updates, without fitting an evolution operator or clustering parameters around representative profiles.}
It maps each direction to a state-dependent output effect and tests whether its realized contributions reconstruct behavior.

\paragraph{Learning dynamics and linearization.}
Infinite-width and lazy-training results describe regimes with approximately fixed linearized dynamics \citep{jacot2018ntk,lee2019wide,chizat2019lazy}.
\rev{Feature-learning, kernel-evolution, and mean-field work instead emphasizes time-dependent representations \citep{woodworth2020kernelrich,yang2021feature,mei2018meanfield,mei2019meanfield,seleznova2022ntkbeyond,loo2022evolution}.}
Deep linear networks also admit exact mode-wise learning dynamics under structured assumptions \citep{saxe2014exact}.
Mode Dynamics analyzes realized finite nonlinear trajectories and directly compares evolving induced effects with a frozen description.

\paragraph{Low-dimensional optimization and adaptation.}
Gradients and training trajectories often occupy low-dimensional subspaces or manifolds \citep{gurari2018tiny,li2018intrinsic,li2021dldr,mao2024manifold}, and low-rank update families are central to efficient adaptation \citep{aghajanyan2021intrinsic,hu2021lora,sonthalia2025lowrank}.
These works motivate compact parameter-space descriptions.
Mode Dynamics asks the complementary question of how trajectory directions contribute to behavior over time.

\paragraph{Behavioral attribution and mechanistic interpretability.}
\rev{Sparse-feature and circuit methods primarily analyze activations and computation within trained models \citep{elhage2022toy,olsson2022induction,wang2022ioi,bricken2023dictionary,templeton2024scaling,gao2024sae,lieberum2024gemmascope,anthropic2025methods,openai2025circuits}.}
Mode Dynamics complements this work by studying how behavior develops over training, linking full-parameter trajectory directions to time-varying output effects and testing whether their contributions reconstruct output change.
\rev{Path-integrated attribution is also used to credit input features along an input interpolation \citep{sundararajan2017axiomatic}, to credit training examples by accumulating per-checkpoint gradient terms \citep{pruthi2020tracin}, and to allocate loss change to individual parameters \citep{lan2019lca}.}
\rev{We instead integrate along the recorded parameter trajectory and attribute output change to collective update directions, for which these methods give no comparable quantity.}
Across these literatures, Mode Dynamics fills the specific gap between trajectory-level update modes and the evolving behavior they produce.

\section{Mode Dynamics Framework}
\label{sec:method}

The framework starts from the ordered parameter states
$\theta_0,\theta_1,\ldots,\theta_T$ and a chosen model output $f(x,\theta)$.
It then follows one dependency chain.
The parameter states define updates, whose decomposition produces modes and amplitudes.
The parameter-to-output Jacobian maps each mode to an induced effect.
Amplitude weighting gives realized contributions that reconstruct output change, and their distribution defines behavioral concentration.

\subsection{Parameter-update decomposition}

Let $\theta_t\in\mathbb{R}^{P}$ be the parameters at checkpoint $t$ and
\begin{equation}
v_t=\theta_{t+1}-\theta_t
\label{eq:update}
\end{equation}
the observed checkpoint-to-checkpoint parameter update.
With checkpoint index treated as discrete time, $v_t$ is the corresponding discrete analogue of velocity.
Stacking the $T$ updates gives
$V=[v_0,\ldots,v_{T-1}]\in\mathbb{R}^{P\times T}$.
Let $r=\operatorname{rank}(V)$.
The thin singular value decomposition (SVD)
$V=U\Sigma W^\top$ produces orthonormal trajectory modes
$u_1,\ldots,u_r$ and their checkpoint-dependent amplitudes
\begin{equation}
a_{k,t}
=u_k^\top v_t
=(\Sigma W^\top)_{k,t}.
\label{eq:amplitude}
\end{equation}
Consequently, every recorded update has the exact decomposition
\begin{equation}
v_t=\sum_{k=1}^{r}a_{k,t}u_k.
\label{eq:decomposition}
\end{equation}
The trajectory directions $u_k$ are fixed for the analyzed trajectory.
Only their amplitudes vary across checkpoints.
The basis is retrospective: it is computed from the complete recorded trajectory rather than used to predict future updates online.
It depends on the parameter coordinates and checkpoint schedule.
Individual modes are non-unique up to sign and rotations within degenerate singular subspaces, as detailed in Appendix~\ref{app:assumptions}.

\subsection{Mode-induced effects and local reconstruction}

The key step is to associate each parameter-space mode with the output effect it induces.
For model output $f(x,\theta)$, let
$J_\theta f(x,\theta)=\partial f(x,\theta)/\partial\theta$
denote the Jacobian of the outputs with respect to the parameters.
The induced effect of a unit movement along mode $u_k$ at state $\theta_t$ is the Jacobian--vector product (JVP)
\begin{equation}
\phi_k(x,\theta_t)
=
J_\theta f(x,\theta_t)u_k.
\label{eq:phi}
\end{equation}
Weighting this unit effect by the observed amplitude defines the realized contribution of mode $k$ as
\begin{equation}
g_{k,t}(x)
=
a_{k,t}\phi_k(x,\theta_t).
\label{eq:contribution}
\end{equation}
Let
$\Delta f_t(x)=f(x,\theta_{t+1})-f(x,\theta_t)$
be the observed one-step output change.
Its local modal reconstruction is
\begin{equation}
\Delta f_t(x)
\simeq
\sum_{k=1}^{r}g_{k,t}(x).
\label{eq:local}
\end{equation}
Equation~\eqref{eq:decomposition} is an exact parameter-space identity, whereas Eq.~\eqref{eq:local} is a first-order approximation in output space.
\rev{Its modal terms are additive but can align or cancel, and the mixed higher-order terms of a finite step are not attributed to individual modes or their interactions.}

\subsection{Accumulated reconstruction and path dependence}

The observed output evolution obeys the exact telescoping identity
\begin{equation}
f(x,\theta_t)
=
f_0(x)+\sum_{\tau<t}\Delta f_\tau(x),
\label{eq:telescoping}
\end{equation}
where $f_0(x)=f(x,\theta_0)$.
Replacing each local change by its modal reconstruction gives
\begin{equation}
\widehat f_t(x)
=
f_0(x)+\sum_{\tau<t}\sum_{k=1}^{r}g_{k,\tau}(x).
\label{eq:global}
\end{equation}
We use path dependence in a specific operational sense.
Although each mode $u_k$ is fixed, its induced effect
$\phi_k(x,\theta_\tau)$ can change with the parameter state visited along the recorded trajectory.
The frozen baseline preserves the same modes and amplitudes but evaluates every induced effect at one reference state.
In all reported frozen baselines, $\theta_{\mathrm{ref}}=\theta_0$, so we call this the \emph{initialization-frozen reconstruction}.
\begin{equation}
\widehat f_t^{\,\mathrm{fixed}}(x)
=
f_0(x)+
\sum_{\tau<t}\sum_{k=1}^{r}
a_{k,\tau}\phi_k(x,\theta_{\mathrm{ref}}).
\label{eq:fixed}
\end{equation}
The evolving and frozen reconstructions therefore differ only in whether
$\phi_k$ is evaluated at $\theta_\tau$ or at $\theta_{\mathrm{ref}}$.
Their contrast tests whether amplitude accumulation with one reference-state Jacobian adequately describes behavior along the observed trajectory.
It does not compare alternative training paths or imply that a fixed endpoint has multiple outputs.

\subsection{Behavioral concentration}
\label{sec:concentration}

Having defined the realized contributions $g_{k,t}$, we ask how their behavioral energy is distributed.
We separately test whether a small ranked subset reproduces output change.

\paragraph{Checkpoint-specific energy.}
The behavioral energy and normalized share of mode $k$ at checkpoint $t$ are
\begin{equation}
E_{k,t}=\|g_{k,t}\|_2^2,
\qquad
\rho_{k,t}=\frac{E_{k,t}}{\sum_j E_{j,t}},
\label{eq:energy}
\end{equation}
\rev{where $j$ runs over all $r$ modes and the squared norm sums over the analysis inputs $x$ and output coordinates stated for each experiment.}
These shares describe the distribution of individual realized-contribution energies at a particular training stage.
Because the contributions need not be orthogonal in function space, the shares are not explained-variance fractions of their sum.

\paragraph{Trajectory-level ordering.}
\rev{Let $\mathcal{S}$ denote the experiment-specific set of checkpoints $t$ whose outgoing updates are used for behavioral ranking.}
To obtain one ordering across these checkpoints, define
\begin{equation}
\begin{aligned}
G_k(x)&=\sum_{t\in\rev{\mathcal{S}}}g_{k,t}(x),\\
\bar E_k&=\|G_k\|_2^2,
\qquad
\bar\rho_k=\frac{\bar E_k}{\sum_j\bar E_j}.
\end{aligned}
\label{eq:trajectory-energy}
\end{equation}
\rev{Here $\bar E_k$ is the energy of the mode's net accumulated contribution, not the time-integrated energy $\sum_{t\in\mathcal{S}}E_{k,t}$, and bars mark trajectory-level quantities rather than averages.}
We call the ordering induced by $\bar\rho_k$ \emph{behavioral-energy ordering}.
We also use \emph{parameter-energy ordering}, which ranks modes by parameter-motion energy $\sigma_k^2=\sum_t a_{k,t}^2$, equivalently the squared norm of the mode's amplitude sequence, and requires no analysis inputs.
Each experiment states its ordering explicitly.

\paragraph{Concentration summaries.}
Let
$\rho_{(1),t}\ge\rho_{(2),t}\ge\cdots$
denote the checkpoint-specific shares in descending order.
The cumulative energy share of the leading $K$ modes is $C_K(t)=\sum_{k=1}^{K}\rho_{(k),t}$.
Our primary threshold-free summary of the full distribution is the participation ratio (PR)
\begin{equation}
\mathrm{PR}(t)
=
\frac{1}{\sum_k\rho_{k,t}^2}.
\label{eq:pr}
\end{equation}
For an intuitive thresholded count, we additionally report
\begin{equation}
K_{\mathrm{eff}}(\tau,t)
:=
\min\left\{K:C_K(t)\ge\tau\right\},
\label{eq:keff}
\end{equation}
using $\tau=0.9$ unless stated otherwise.
The same cumulative, PR, and effective-count summaries apply to the trajectory-level shares $\bar\rho_k$.

\paragraph{Top-\texorpdfstring{$K$}{K} reconstruction fidelity.}
For a stated ordering, Top-$K$ reconstruction retains the $K$ highest-ranked modes and measures how well they reproduce output change using metrics such as $R^2$, reconstruction error, KL divergence, or prediction agreement.
Rapid fidelity improvement with $K$ supports behavioral compactness, but fidelity is distinct from energy concentration because the realized contributions $g_{k,t}$ can align or cancel in function space.
PR and $K_{\mathrm{eff}}$ require the relevant full energy distribution, whereas Top-$K$ fidelity can also be evaluated for a truncated set of leading modes.

\subsection{Computing mode-induced effects}

Equation~\eqref{eq:phi} can be evaluated without constructing the full Jacobian, using either automatic differentiation (AD) to compute JVPs or directional finite differences.
Regression and controlled next-token prediction use finite differences, whereas CIFAR-100 and Pythia use forward-mode AD and quadrature along the straight \emph{checkpoint chord} joining consecutive checkpoints.
Appendix~\ref{app:assumptions} gives estimator boundaries.
The dominant costs are $O(TP)$ trajectory storage and $O(KT)$ direction--state evaluations, multiplied by $Q$ when each checkpoint chord uses $Q$ quadrature nodes. Appendix~\ref{app:complexity} gives the full accounting.

\section{Experiments}
\label{sec:experiments}

We evaluate the framework in four settings: scalar regression, controlled next-token prediction, CIFAR-100 autoencoding, and a public Pythia-70M pretraining trajectory.
All four test whether leading trajectory-derived modes reconstruct observed output change.
The two controlled experiments provide detailed mechanism and control tests.
CIFAR-100 and Pythia-70M provide compact transfer tests across image and language outputs.
Appendix Table~\ref{tab:overview} summarizes the complementary tasks and metric coverage, and the appendix gives complete protocols.

For every reported reconstruction $R^2$, let $s$ denote the reference model output or output change, stacked over the stated examples and output coordinates, and let $\widehat s$ denote its modal reconstruction.
We report
\begin{equation}
R^2=1-\frac{\|\widehat s-s\|_2^2}{\|s-\overline{s}\|_2^2}.
\label{eq:r2}
\end{equation}
Here $\overline{s}$ is the mean of the entries of $s$.
Values below zero indicate reconstruction error greater than that of the constant baseline $\overline{s}$.
The reference is always the model output being reconstructed, not the task target.
Each experiment states whether this reference is a one-step output change, an accumulated change from initialization, or a model output at a specified checkpoint.
For Pythia, the same ratio is applied to vocabulary-centered logit change pooled over contexts, as specified in Appendix~\ref{app:pythia}.

\subsection{Controlled Mechanistic Validation}
\label{sec:controlled}

\subsubsection{Scalar Regression}
\label{sec:regression}

Regression tests local fidelity and temporal accumulation in a setting where prediction curves and residuals can be inspected directly.
We fit $y=\sin x$ on $[0,2\pi]$ with a one-hidden-layer multilayer perceptron (MLP) using mean-squared error (MSE) and stochastic gradient descent (SGD).
The main grid contains 15 trajectories, with five seeds at each width in $\{32,64,128\}$.
\rev{We compare against three baselines.}
\rev{Dimension-matched random subspaces test whether trajectory-derived directions matter, and the initialization-frozen reconstruction tests path dependence.}
\rev{An empirical neural tangent kernel (NTK) trains the network linearized at initialization, so it cannot capture a changing Jacobian.}
Appendix~\ref{app:regression} gives optimization, checkpoint, robustness, and control details.

\paragraph{Local reconstruction.}

\begin{figure}[t]
\centering
\includegraphics[width=0.84\linewidth]{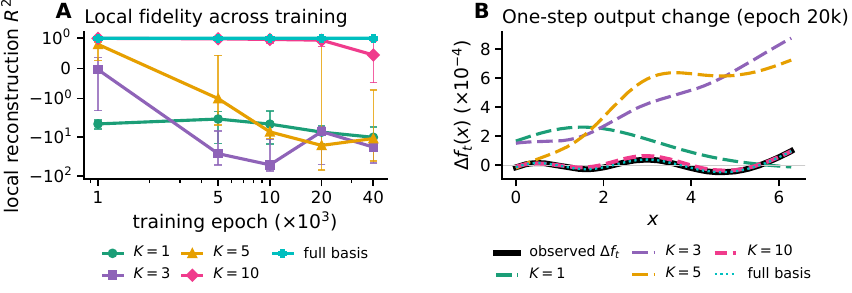}
\caption{\textbf{Leading-mode reconstructions approximate local output changes.}
\textbf{A:} Median local $R^2$ for $K\in\{1,3,5,10\}$ and the full basis over 15 trajectories.
Capped bars show the interquartile range and the vertical axis is symmetric-log.
\textbf{B:} The observed and reconstructed one-step output change for a representative run at epoch 20k.
Modes follow the parameter-energy ordering of the update SVD.}
\label{fig:local-decomposition}
\end{figure}

Figure~\ref{fig:local-decomposition} shows that a single mode is insufficient, whereas Top-10 approaches the full-basis reconstruction at early and intermediate checkpoints.
For the reference width-64 architecture, Top-10 uses only $10/193\approx5.2\%$ of the available trajectory modes.
Near convergence, one-step output changes have negligible variance, so tiny absolute errors can make local Top-$K$ $R^2$ unstable.
Full-basis local fidelity and accumulated Top-$K$ reconstruction nevertheless remain high.

\paragraph{Path-dependent induced effects.}

\begin{figure}[t]
\centering
\includegraphics[width=0.84\linewidth]{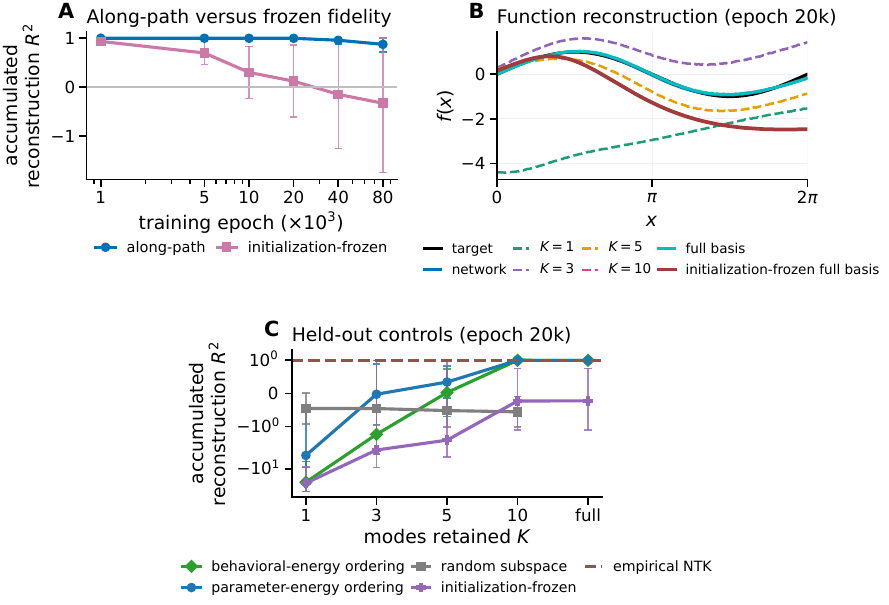}
\caption{\textbf{Path-dependent induced effects improve accumulated reconstruction.}
\textbf{A:} Mean Top-10 accumulated reconstruction $R^2$ over 15 runs for along-path and initialization-frozen induced effects.
\textbf{B:} Representative along-path Top-$K$ function reconstruction at epoch 20k with the initialization-frozen full-basis baseline.
\textbf{C:} Controls at epoch 20k over five width-64 seeds, scored on a separate analysis set held out from behavioral-energy ranking.
Capped bars show one standard deviation.
Panel A uses logarithmic epoch spacing, Panel C uses symmetric-log vertical scaling, and Panels A--B use trajectory-level behavioral-energy ordering.
Protocols and control definitions are in Appendix~\ref{app:regression}.}
\label{fig:global-reconstruction}
\end{figure}

Figure~\ref{fig:global-reconstruction} shows that Top-10 reconstruction is nearly exact through 20k, then becomes less accurate and more variable at later checkpoints.
The initialization-frozen baseline degrades progressively.
Behavioral-energy and parameter-energy orderings retain high Top-10 accumulated $R^2$, whereas dimension-matched random subspaces perform poorly on average.
The empirical NTK also remains predictive of the regression function, while Mode Dynamics attributes output change along the realized finite-network trajectory.
Reconstruction remains strong across seeds, widths, and checkpoint schedules in the SGD grid (Appendix Figure~\ref{fig:supp:regression-robustness}).

\paragraph{Behavioral concentration.}

The checkpoint-by-checkpoint analysis in Appendix Figure~\ref{fig:supp:regression-dynamics} complements the trajectory-level reconstruction ordering by showing how checkpoint-specific behavioral energy is distributed during learning.
PR remains small throughout training.
Separately, trajectory-derived orderings reach high fidelity by $K=10$, while random-subspace and initialization-frozen controls remain poor (Figure~\ref{fig:global-reconstruction}C).
Across the 15 trajectories, Top-10 accumulated $R^2$ at epoch 20k is $0.998\pm0.002$.
At epoch 10k, the checkpoint-specific mean ranked realized-contribution energy profile has $\mathrm{PR}=3.7$ and $K_{\mathrm{eff}}(0.9)\approx4$.

\subsubsection{Controlled Next-Token Prediction}
\label{sec:transformers}

We use small decoder-only transformers on three controlled last-position token-prediction tasks.
The tasks are \emph{copying}, which predicts the visible final content token, \emph{induction}, which retrieves the token that followed a repeated anchor, and \emph{random-label memorization}, which assigns one fixed random content-token label to each sequence.
Let $L$ denote sequence length and $V_{\mathrm{c}}$ content-vocabulary size.
The systematic grid crosses five seeds, depths $\{1,2,3\}$, and 15 valid $(L,V_{\mathrm{c}})$ pairs, yielding 225 runs per task and 675 total.
The pairs are drawn from lengths $\{4,6,8,10,12\}$ and vocabularies $\{6,8,12,16\}$ subject to $V_{\mathrm{c}}\ge L-1$.
All models reach perfect training accuracy.
Top-$K$ logit reconstructions are evaluated across five seeds in a reference configuration (depth 1, sequence length 4, content vocabulary 6).
\rev{Except in the causal test below, no train--test split is used, so all metrics are in-sample on the finite training sequences, and seed variation includes the generated sequences or labels.}
For predicted-token decisions, we report \emph{Top-1 prediction agreement}, the fraction of examples for which the trained model and Top-$K$ reconstruction have the same highest-logit token.
Architecture, optimization, sequence construction, and estimator details are in Appendix~\ref{app:transformers}.

\paragraph{Concentration across task and data conditions.}

\begin{figure}[t]
\centering
\includegraphics[width=0.84\linewidth]{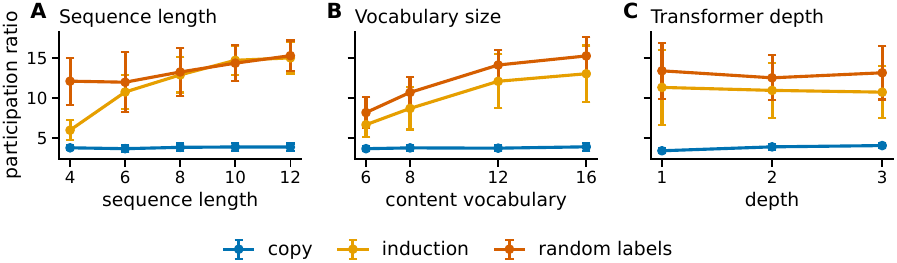}
\caption{\textbf{Behavioral PR varies across next-token tasks and data conditions.}
At epoch 50k, the full 675-run grid shows participation ratio versus sequence length in Panel A, content vocabulary in Panel B, and transformer depth in Panel C.
Error bars denote one standard deviation over the corresponding conditions and seeds.
All metrics are evaluated on training sequences.
Matched-condition optimization, accuracy, and PR trajectories are in Appendix~\ref{app:transformers}.}
\label{fig:transformer-dynamics}
\end{figure}

Figure~\ref{fig:transformer-dynamics} summarizes the epoch-50k grid, while matched-condition trajectories in Appendix Figure~\ref{fig:supp:transformer-training} show that copying remains compact as induction and random-label memorization recruit additional modes during training.
Across the full grid at epoch 50k, pooled PR is $3.8\pm0.4$ for copying, $11.0\pm3.9$ for induction, and $13.0\pm3.2$ for random labels.
Behavioral PR is therefore associated with task identity even when all models ultimately fit their training sets.
In the marginal summaries over valid pairs, larger sequence length or vocabulary is associated with higher PR for induction and random-label memorization, while copying remains nearly unchanged (Panels A--B).
Because $V_{\mathrm{c}}\ge L-1$ couples the available sequence and vocabulary conditions, these panels are descriptive marginal associations rather than isolated causal effects.
Restricted-support checks in Appendix~\ref{app:transformers} preserve the task ordering under this coupling.
Within the tested grid, mean PR varies less across depths than across tasks despite increasing parameter count (Panel C).
This does not establish scale invariance.
Appendix~\ref{app:dynamic-static} compares trajectory dimension with static endpoint rank.

\paragraph{Behavioral reconstruction.}

\begin{figure}[t]
\centering
\includegraphics[width=0.84\linewidth,trim=0 118.3 0 0,clip]{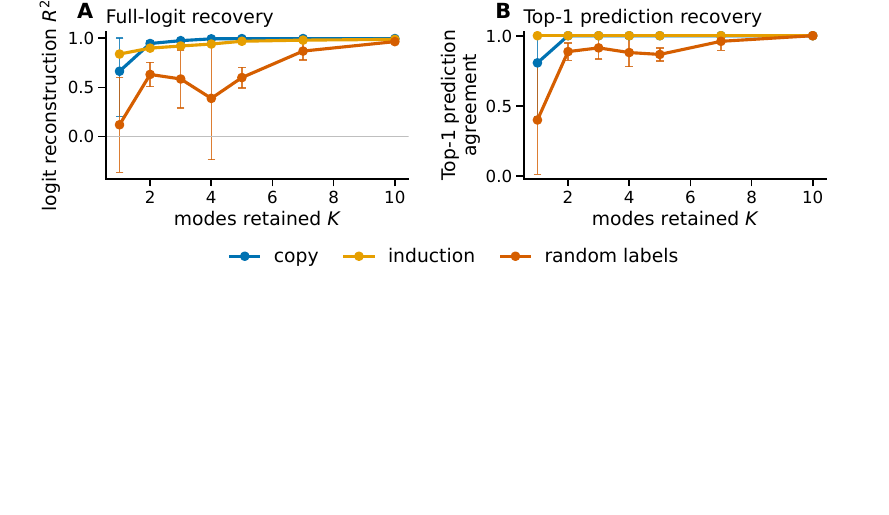}\\[4pt]
\includegraphics[width=0.84\linewidth]{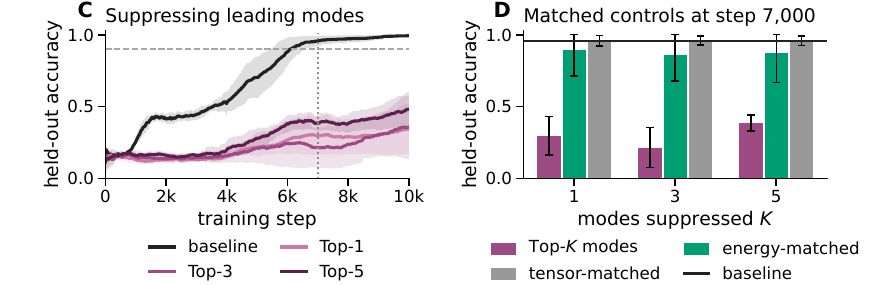}
\caption{\rev{\textbf{Small sets of leading modes reconstruct next-token training-set behavior, and suppressing them during training impairs held-out induction accuracy.}}
\textbf{A--B:} Training-set logit $R^2$ and Top-1 prediction agreement compare each Top-$K$ reconstruction with the trained full-network output, averaged over five seeds.
Capped bars show one population standard deviation, and modes use trajectory-level behavioral-energy ordering.
\rev{\textbf{C--D:} Each run restarts from a reference run's initialization and projects every gradient away from a fixed set of that run's modes.
Accuracy is measured on 216 held-out induction contexts, with mean $\pm$ one population standard deviation over five seeds.
\textbf{C:} Learning curves under Top-$K$ suppression.
The dashed line marks $90\%$ accuracy and the dotted line marks step 7,000.
\textbf{D:} Accuracy at step 7,000 when suppressing the Top-$K$ modes, $K$ energy-matched modes, or the tensor-matched Top-$K$ modes, with the unsuppressed baseline as a horizontal line.
The protocol is in Appendix~\ref{app:causal}.}}
\label{fig:transformer-reconstruction}
\end{figure}

Figure~\ref{fig:transformer-reconstruction} links concentration to behavioral fidelity without equating energy and accuracy.
Across the shared $K=1,\ldots,10$ sweep, copying reaches 100\% Top-1 prediction agreement by $K=2$ and induction by $K=1$, while random labels approach consistent recovery only near $K=10$.
Additional modes continue to improve the numerical accuracy of reconstructed logits, especially for induction.
\rev{In the selected examples of Appendix Figure~\ref{fig:supp:reconstruction-examples}, Panels C and E visibly correct a low-$K$ decision, whereas Panel D preserves induction's already-correct decision as its logits become more faithful.}
Appendix~\ref{app:heatmaps} repeats the same cumulative row-wise view for three additional selected examples per task.

\begin{revblock}
\paragraph{Causal suppression.}
Reconstruction alone does not show that learning depends on the leading modes, so we suppress them during training (Appendix~\ref{app:causal}).
For each of five seeds, a depth-1 transformer is trained on 864 of the 1,080 valid induction contexts with $L=V_{\mathrm{c}}=6$, and its modes are ranked on these contexts only.
A second run restarts from the same initialization and projects every gradient away from the Top-$K$ modes, $K\in\{1,3,5\}$, so the parameters never move along them.
Energy-matched controls instead suppress $K$ modes with similar parameter-motion energy $\sigma_k^2$ but smaller output effect.
Tensor-matched controls suppress the Top-$K$ modes after one random signed permutation within each parameter tensor, which keeps each mode's length but changes which parameters move together.
On the 216 held-out contexts, step-7,000 accuracy falls from $95.7\%$ to $29.8\%$, $21.5\%$, and $38.6\%$ under Top-1, Top-3, and Top-5 suppression, versus $86$--$89\%$ for energy-matched and about $96\%$ for tensor-matched controls (Figure~\ref{fig:transformer-reconstruction}C--D).
No Top-$K$-suppressed run reaches $90\%$ by step 10,000, whereas all baseline and tensor-matched runs and 13 of 15 energy-matched runs do.
Because the modes come from a completed reference run, this test is reference-informed rather than prospective.

\paragraph{Initialization and checkpoint density.}
In the grid's depth-1 induction condition with $L=V_{\mathrm{c}}=6$ and 30 training sequences, five runs that differ only in initialization give trajectory-level PR $1.99\pm0.52$ and $K_{\mathrm{eff}}(0.9)$ from 2 to 5 (Appendix~\ref{app:robustness}).
Refining checkpoint spacing from 1,000 to 100 steps raises mean $K_{\mathrm{eff}}(0.9)$ only from 2.0 to 3.0, and the five leading modes reconstruct the net all-mode contribution with $R^2>0.96$ in every run at every spacing, although the leading subspaces shift.
\end{revblock}

\FloatBarrier
\subsection{Validation on CIFAR-100 and Pythia-70M}
\label{sec:natural-validation}

The controlled settings isolate reconstruction, path dependence, and behavioral concentration.
We next evaluate the framework on held-out CIFAR-100 images and a public Pythia-70M pretraining trajectory.
Table~\ref{tab:natural-validation} summarizes the main reconstruction, concentration, decision, and control results.
We summarize the protocols here and provide full details in Appendices~\ref{app:cifar} and \ref{app:pythia}.
For these checkpointed systems, discrete parameter updates define the trajectory SVD, while each pointwise response term in Equation~\eqref{eq:global} is replaced by a response integral along each checkpoint chord, estimated by quadrature.

\begin{table}[!ht]
\centering
\caption{\textbf{Summary of CIFAR-100 and Pythia-70M results.}
CIFAR ``$\pm$'' values are population standard deviations over five trajectories, except the control, which is over three seed-0 signed permutations.
Pythia rows above the \emph{Post-hoc sensitivity} break use analysis settings fixed before any Pile-test access. Those below vary checkpoint density and quadrature afterwards on the same frozen test contexts.
The bracketed interval is a 95\% context-bootstrap confidence interval.}
\label{tab:natural-validation}
{\footnotesize
\setlength{\tabcolsep}{3pt}
\renewcommand{\arraystretch}{1.02}
\resizebox{\linewidth}{!}{%
\begin{tabular}{@{}llll@{}}
\toprule
\textbf{Analysis} & \textbf{Retained set / control} & \textbf{Metric} & \textbf{Result} \\
\midrule
\multicolumn{4}{@{}l}{\textbf{CIFAR-100 autoencoding}} \\
\cmidrule(l){1-4}
Test reconstruction & Top-50 & pooled RGB-change $R^2$ & $0.912\pm0.012$ \\
Image-level fidelity & Top-50 & median per-image $R^2$ & $0.892$ \\
Parameter-motion energy & Top-50 & fraction retained & $43.8\%\pm0.4\%$ \\
Coordinate control (training split) & signed permutation & pooled RGB-change $R^2$ & $-0.642\pm0.344$ \\
\addlinespace[3pt]
\multicolumn{4}{@{}l}{\textbf{Pythia-70M next-token pretraining}} \\
\cmidrule(l){1-4}
Held-out reconstruction & 20/50 modes (40\%), $Q=2$ & pooled centered $\Delta$logit $R^2$, Top-1 prediction agreement & $0.9668\ [0.9662,0.9673]$, $0\%$ \\
Validation concentration & 51 checkpoints / 50 modes & PR, $K_{\rm eff}(0.9)$ & $5.09$, $13$ \\
Coordinate / state controls & signed-perm. 50-mode path, init.-frozen all-mode & pooled centered $\Delta$logit $R^2$ & $-0.890,\ -0.018$ \\
\addlinespace[2pt]
\multicolumn{4}{@{}l}{\emph{Post-hoc sensitivity}} \\
Full-update convergence & 51 ckpt, no modal projection & Top-1 prediction agreement & $77.3\%$ at $Q=2$ (locked), $97.7\%$ at $Q=8$ \\
Distribution / decision recovery & 101 ckpt, 80/100 modes (80\%), $Q=8$ & mean KL (nats), Top-1 prediction agreement & $0.531$, $64.1\%$ \\
Full-update reference & 101 ckpt, no modal projection, $Q=8$ & mean KL (nats), Top-1 prediction agreement & $0.00425$, $96.9\%$ \\
\bottomrule
\end{tabular}}}
\end{table}

\paragraph{CIFAR-100 autoencoding.}
We train five convolutional autoencoders on CIFAR-100 \citep{krizhevsky2009learning} and record 1,961 checkpoints per trajectory.
For each trajectory we estimate the leading 50 update-SVD modes in the parameter-energy ordering.
The \emph{locked} evaluation was fixed before any test score was inspected.
It uses the path-integrated contributions of these modes to reconstruct the decoder's RGB output change from initialization on 100 test images per seed, held out from training, modal scoring, and mode selection (Appendix~\ref{app:cifar}).
The 50-mode reconstruction attains pooled RGB-change $R^2=0.912$ while retaining under half of the parameter-motion energy (Table~\ref{tab:natural-validation} and Appendix Figures~\ref{fig:cifar-rgb}--\ref{fig:supp:cifar-images}).
\rev{In a separate training-split analysis, fixed signed permutations randomly reassign and sign-flip parameter coordinates, which keeps each mode's length and the angles between modes but changes each mode's direction.}
Their strongly negative result shows that recovering output change requires preserving which parameters carry the coordinated motion, not update geometry alone.
\rev{A post-hoc tensor-matched control, which restricts signed permutations to each RGB-relevant tensor and leaves the classifier head unchanged, also fails on the held-out images, with $R^2=-0.909\pm0.559$ (Appendix~\ref{app:cifar-tensor}).}
Because only a truncated 50-mode basis is estimated, this experiment supports compact reconstruction but does not identify a full-basis behavioral PR.

\paragraph{Pythia-70M next-token pretraining.}
We analyze 51 released Pythia-70M checkpoints spanning pretraining steps 0--143k, yielding 50 checkpoint updates and a corresponding full numerical mode basis.
The reconstructed output is the final-position logit change from initialization, centered by subtracting each context's mean logit across the vocabulary.
The analysis is likewise locked, with every setting fixed before any Pile-test access (Appendix~\ref{app:pythia}).
Modes are ranked once on 128 Pile-validation contexts and scored without reranking on 128 untouched Pile-test contexts.
Using two quadrature nodes per checkpoint chord ($Q=2$), the leading 20 modes attain held-out pooled centered $\Delta$logit $R^2=0.9668$, with validation behavioral energy PR $5.09$ and $K_{\mathrm{eff}}(0.9)=13$ (Appendix Figure~\ref{fig:supp:pythia-reconstruction}).
\rev{In a post-hoc comparison at the same budget, the five validation-ranked modes with the largest output effect give $R^2=0.728$, versus $0.134$ for the five with the largest parameter motion, although behavioral-energy ordering does not uniformly improve KL divergence or Top-1 prediction agreement (Appendix Table~\ref{tab:pythia-ordering}).}
The signed-permutation and initialization-frozen controls both fail on the same held-out contexts (Table~\ref{tab:natural-validation}), indicating that coordinate identity and along-path evaluation both matter for reconstruction in this setting.
Despite this high geometric fidelity, the 20-mode reconstruction has zero Top-1 prediction agreement.
\rev{Pooled $R^2$ weights all 50,304 logits, whereas a prediction depends only on the largest, so $R^2$ measures output-change reconstruction, not decision recovery.}
Two \emph{post-hoc} refinements probe this decision gap by changing analysis settings after the locked result and rescoring the same frozen test contexts (Appendix Table~\ref{tab:pythia-sensitivity}).
Increasing quadrature order makes the full-update reference converge toward the endpoint change.
Because this reference integrates complete checkpoint updates without modal projection and therefore realizes Equation~\eqref{eq:telescoping} directly, its residual error is numerical rather than modal.
The nested 101-checkpoint schedule supplies more modes to retain: its 80-mode reconstruction reaches $64.1\%$ agreement at mean Kullback--Leibler (KL) divergence $0.531$ nats (natural-log units), compared with $96.9\%$ agreement and $0.00425$ nats for the full-update reference.
At 80\% retention, the remaining decision gap therefore reflects truncation rather than quadrature error.
Relaxing the rank cutoff, the trained model's predicted token lies within that reconstruction's ten highest-logit candidates for $88.3\%$ of contexts, while the signed-permutation control never reaches the top fifty of a 50,304-token vocabulary (Appendix Figure~\ref{fig:supp:pythia-checkpoint-quadrature}).
We interpret this single public trajectory as evidence for compact centered logit-change geometry, not decision-preserving model compression.

\FloatBarrier
\section{Discussion}
\label{sec:discussion}

\rev{Across all four settings, the results support one account of learning: decomposition finds collective modes of parameter velocity, reconstruction shows that small subsets of these modes recover most of the output change, and concentration measures how that change is distributed across modes.}
\rev{Modes that carry the most output change need not be those with the most parameter motion.}
\rev{In a controlled induction task, suppressing them during training impairs held-out accuracy, whereas matched controls do not.}
Along-path effects outperform initialization-frozen effects in the tested comparisons.
Trajectory-derived directions outperform dimension-matched random subspaces, and signed-permutation controls fail to reconstruct output change.
In the controlled experiments, behavioral concentration varies systematically with task and data, while in the transfer settings it remains only partially characterized.

\paragraph{Implications.}
Trajectory modes provide a parameter-space account of how outputs change during learning, complementing activation-level methods that study computation in a fixed trained network.
Reconstruction attributes output change to coordinated parameter directions, providing a basis for monitoring behavior changes and auditing the associated parameter motion.
Concentration distinguishes changes carried by a few directions from broadly distributed changes, which may inform hypotheses about behavioral fragility or robustness.
\rev{Safety and alignment are natural settings for these questions \citep{barez2026safety}, but demonstrating such uses requires further causal evidence.}

\paragraph{Scope and limitations.}
The method requires checkpoints containing the same complete set of parameter tensors in a consistent coordinate order.
Computing mode-induced effects also requires JVPs through the model, making the method heavier than analyses based only on final weights.
The framework quantifies concentration without explaining why it arises.
\rev{It does not localize modes to model components or circuits, and its causal evidence is limited to suppressing reference-run modes in one controlled induction task within a fixed training budget.}
\rev{Because each run has its own basis, we have not tested whether the same modes recur or activate in the same order across runs.}
The transfer analyses have different basis limitations: the image-autoencoder basis is truncated, while the language-model mode count is limited by recorded intervals and is therefore resolution-dependent.
\rev{Counts also grow modestly with checkpoint density in the controlled induction task, so across settings we claim compactness over the tested schedules, not converged counts.}
Evidence is limited to one 730k-parameter autoencoder architecture and one 70M-parameter language-model pretraining trajectory, so generalization across architectures and scales remains unresolved.
The language-model results also show that near-perfect centered-logit reconstruction can still leave a decision gap.

\paragraph{Future work.}
Questions of resolution and scaling are computational rather than conceptual.
\rev{Denser checkpoint schedules for language models would test whether their counts stabilize, while broader architectures and larger models would test generality.}
Online variants that update the mode basis as training proceeds would be needed for real-time monitoring.
\rev{Estimating modes early in training would test whether they can predict later behavior and guide training actions such as hyperparameter changes.}
Relating modes to layerwise activations and localized circuits would connect training dynamics to computation in the trained network.
\rev{Linking modes to downstream capabilities would be a step toward transferring them across models.}

\FloatBarrier
\subsubsection*{Ethics statement}
The work analyzes controlled regression and next-token tasks, CIFAR-100, and public Pile validation and test text using released Pythia checkpoints.
It involves no human-subjects research or model deployment.
The resulting trajectory attributions do not guarantee causal control or model safety.

\subsubsection*{Reproducibility statement}
The Appendix documents the experimental and analysis protocols, numerical methods, and controls for each setting.
Upon acceptance, we will release the code and supporting materials needed to reproduce the reported results.
Bitwise reproducibility across hardware backends is not claimed.

\subsubsection*{AI use statement}
Generative AI assisted with implementing experiment code, scripting figures, and drafting and editing prose.
The authors developed the framework and designed the experiments, controls, and frozen protocols.
They reviewed and revised all AI-assisted text and code.
They ran every experiment, performed sanity and consistency checks, verified every reported number against saved artifacts, and take responsibility for the final content.

\bibliographystyle{iclr2027_conference}
\bibliography{references}

\clearpage
\appendix
\section*{Supplementary Material}

Table~\ref{tab:overview} summarizes the experimental settings and their metric coverage.

\begin{table}[h]
\centering
\caption{\textbf{Experimental overview and metric coverage.}}
\label{tab:overview}
{\footnotesize
\setlength{\tabcolsep}{3pt}
\resizebox{\linewidth}{!}{\begin{tabular}{lllll}
\toprule
setting & task / model & reconstructed output & mode ordering & reported measures \\
\midrule
Scalar regression & \shortstack[l]{sine fitting\\one-layer MLP} & scalar prediction curve & \shortstack[l]{parameter-energy (local)\\behavioral-energy (accumulated)} & \shortstack[l]{local/accumulated Top-$K$ $R^2$\\PR and auxiliary $K_{\rm eff}$} \\
\midrule
Next-token prediction & \shortstack[l]{copy / induction / random labels\\decoder-only transformer} & \shortstack[l]{last-position logits\\and predicted-token decisions} & behavioral-energy & \shortstack[l]{Top-$K$ logit $R^2$ and Top-1 prediction agreement\\PR and auxiliary $K_{\rm eff}$} \\
\midrule
Pixel-space autoencoding & \shortstack[l]{CIFAR-100 reconstruction\\convolutional autoencoder} & RGB decoder-output change & parameter-energy & \shortstack[l]{Top-50 RGB-change $R^2$\\retained parameter-motion energy, no full PR} \\
\midrule
Public pretraining trajectory & \shortstack[l]{Pythia-70M\\validation ranking / locked Pile-test scoring} & \shortstack[l]{vocabulary-centered final-position\\next-token logit change} & \shortstack[l]{behavioral-energy on\\dedicated ranking contexts} & \shortstack[l]{held-out pooled $\Delta$logit $R^2$, PR, $K_{\rm eff}$\\post-hoc KL and Top-$r$ prediction agreement} \\
\bottomrule
\end{tabular}
}}
\end{table}

\section{Methodological Scope}
\label{app:assumptions}

\paragraph{Trajectory scope and mode definition.}
Mode Dynamics retrospectively analyzes a finite recorded trajectory in a fixed parameter coordinate system.
It assumes neither infinite width, Gaussian initialization, nor an NTK regime.
The complete trajectory defines the regression, controlled next-token, and Pythia SVD bases, while behavioral-energy rankings also depend on the stated analysis inputs and intervals.
These full SVDs reconstruct every recorded parameter update up to numerical precision.
Individual directions remain coordinate-dependent, arbitrary in sign, and defined only up to rotations within degenerate singular subspaces.
Near-degeneracy can destabilize individual directions while preserving their joint span.
The CIFAR-100 analysis instead estimates only the leading 50 modes with randomized truncated SVD.

\paragraph{Exactness and numerical estimation.}
The main-text JVP maps modes to output space and requires the output map to be differentiable almost everywhere along the chosen interpolation.
Unlike full-SVD update reconstruction, local output reconstruction is first-order.
Regression and controlled next-token JVPs use forward finite differences without an $\epsilon$ sweep.
For regression, the along-path estimator applies one finite difference to each aggregate projected update, whereas the initialization-frozen estimator sums modewise finite differences at $\theta_0$.
As $\epsilon\to0$, the estimators converge to state-specific JVPs, with along-path evaluation at the current states and frozen evaluation at initialization.
Their contrast therefore tests the stated frozen baseline, while finite $\epsilon$ adds estimator approximation.
CIFAR-100 uses forward-mode AD along straight segments between consecutive checkpoints, avoiding finite-difference error but retaining quadrature, floating-point, nonsmooth-derivative, checkpoint-spacing, and interpolation dependencies.
Pythia uses the same chord-wise forward-mode construction with two-point Gauss--Legendre quadrature and vocabulary-centered final-position logit change in the locked analysis.
Checkpoint density and quadrature order are distinct.
Additional saved checkpoints change the update matrix and mode basis, shorten the piecewise-linear chords, and can change truncated projections.
Additional quadrature nodes leave the checkpoints and basis fixed and only improve integration along those chords.
For Pythia, \emph{full-update} means integrating complete checkpoint updates without any modal projection, while \emph{all-mode} means summing the full numerical mode basis.
The full-update Jacobian line integral is an exact differential and therefore recovers the endpoint output change for any connecting chord in the exact-integration limit.
Residual full-update error diagnoses quadrature or floating-point error.
An all-mode modal reconstruction can additionally inherit numerical SVD and projection error.
Truncated modal attributions, unlike the full sum, can depend on the chosen chord path and checkpoint schedule.

\paragraph{Uncertainty reporting.}
Unless stated otherwise, error bars, bands, and ``$\pm$'' values are descriptive population standard deviations over the runs identified in each caption, not confidence intervals.
Pythia confidence intervals instead use 2,000 context-bootstrap resamples of its locked confirmatory Pile-test contexts.

\section{Computational Complexity}
\label{app:complexity}

For $T$ updates in $P$ parameters, explicitly storing the trajectory requires $O(TP)$ memory.
A dense SVD costs $O(\min\{T^2P,TP^2\})$ in the worst case.
Evaluating $K$ directions at $T$ checkpoints by finite differences requires $O(KT)$ additional forward evaluations on the selected examples.
With $Q$ quadrature nodes per checkpoint chord, path-integrated evaluation requires $O(KTQ)$ direction--node evaluations.
AD JVPs have the analogous direction--state evaluation burden and can be batched.
Leading-mode truncation, randomized or streaming SVD, checkpoint subsampling, minibatch energy estimates, and layerwise decompositions can reduce these costs.

\section{Scalar Regression}
\label{app:regression}

\subsection{Protocol and Reproducibility}

\paragraph{Data and training.}
The reference experiment draws 5,000 inputs independently and uniformly from $[0,2\pi]$ with targets $y=\sin x$.
A one-hidden-layer width-64 tanh MLP (193 parameters) is trained from fan-in-scaled uniform initialization for 80k epochs using full-batch SGD on MSE at learning rate $0.01$, with checkpoints every 10 epochs.
The robustness grid spans widths 32, 64, and 128 with five seeds each (15 trajectories).
Each seed controls both the input draw and parameter initialization.

\paragraph{Evaluation and controls.}
A deterministic 400-point analysis grid on $[0,2\pi]$ is alternated into 200 ranking points and 200 scoring points held out from behavioral-energy ranking.
This is a ranking--scoring analysis split, not a task train--test split.
For each width-64 trajectory, the random baseline uses 20 nested, dimension-matched random subspaces formed by orthonormalizing Gaussian parameter directions.
This construction samples orientations without favoring any parameter coordinate or learned update direction.
Because these subspaces stop at dimension 20, the random control has no full-basis entry and is plotted only up to that budget. A dimension-193 random subspace would span parameter space and make full-basis reconstruction trivial.
The initialization-frozen baseline evaluates induced effects at initialization, while the empirical NTK baseline simulates full-batch gradient descent in the initialization-linearized finite-width model at the same learning rate.
These controls test generic low-dimensional projection and the chosen initialization linearization.
Without cross-trajectory mode swaps, they do not establish that individual modes are trajectory-specific.
Parameter-energy ordering is not an independent random control because integrated squared amplitudes equal squared singular values.

\paragraph{Analysis schedule.}
Induced effects use forward finite differences with $\epsilon=0.001$.
Behavioral energy and local reconstruction are evaluated at 1k, 5k, 10k, 20k, and 40k.
Accumulated reconstructions continue through the terminal 80k checkpoint.

\subsection{Optimization, Concentration, and Robustness}

\begin{figure}[h]
\centering
\includegraphics[width=0.82\linewidth]{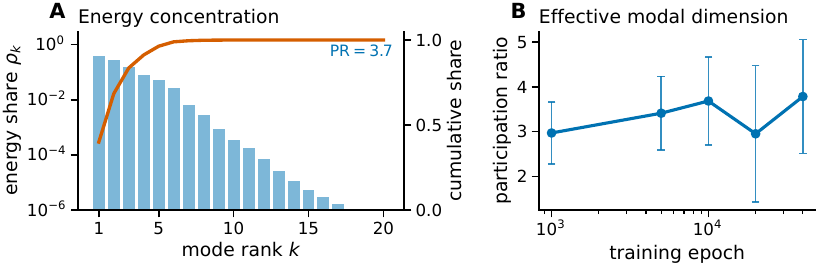}
\caption{\textbf{Regression output-change energy is concentrated across training.}
\textbf{A:} Mean ranked behavioral energy at epoch 10k, its cumulative share, and the corresponding participation ratio.
\textbf{B:} Mean participation ratio across the five analyzed nonterminal checkpoints through 40k, with capped one-standard-deviation bars.
Panel A averages the same five-seed, three-width sine grid.}
\label{fig:supp:regression-dynamics}
\end{figure}

\begin{figure}[h]
\centering
\includegraphics[width=0.58\linewidth]{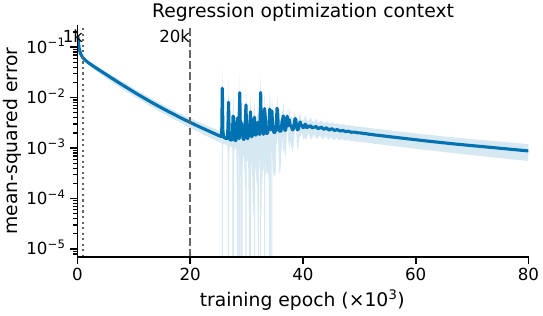}
\caption{\textbf{Regression optimization context.}
Mean width-64 sine-target MSE through 80k.
Shading shows one standard deviation over five seeds.
Vertical markers locate the representative 1k and 20k analysis checkpoints.}
\label{fig:supp:regression-loss}
\end{figure}

\begin{figure}[h]
\centering
\includegraphics[width=0.90\linewidth]{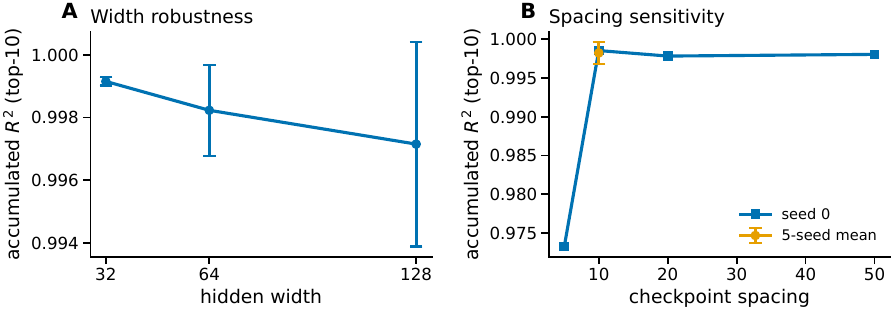}
\caption{\textbf{Seed, width, and checkpoint-spacing robustness.}
\textbf{A:} Sine-target Top-10 accumulated $R^2$ at 20k, aggregated over five seeds at each width.
\textbf{B:} Sine Top-10 accumulated $R^2$ at 20k versus checkpoint spacing for width 64.
Top-10 uses trajectory-level behavioral-energy ordering.
The default spacing of 10 averages five seeds, while additional spacing runs use seed 0 only.}
\label{fig:supp:regression-robustness}
\end{figure}

\FloatBarrier
The loss decreases monotonically through 25k and then rises on $22\%$ of logged steps between 25k and 40k, the same interval over which Top-10 accumulated reconstruction degrades and its across-seed spread widens.
Figure~\ref{fig:supp:regression-robustness} shows that high-fidelity Top-10 accumulated reconstruction is reproducible across seeds and widths.
Panel B is a descriptive spacing check.
Changing spacing also changes the update matrix, SVD basis, and temporal approximation, so this comparison does not isolate discretization error.
As a compact target-function check, the same 15-run grid fitted
$\sin x+0.5\sin 2x+0.25\sin 3x$ and achieved Top-10 accumulated
$R^2>0.99$ at epoch 20k.

\FloatBarrier
\section{Controlled Next-Token Prediction}
\label{app:transformers}

\subsection{Protocol and Reproducibility}

\paragraph{Model and tasks.}
The decoder-only transformers use hidden dimension 32, two attention heads, learned token and position embeddings, and a linear output head.
Each block contains pre-normalized causal self-attention followed by a pre-normalized fourfold-expansion GELU MLP, with no dropout.
Depth ranges from one to three.
For sequence length $L$ and content-vocabulary size $V_{\mathrm{c}}$, four task tokens are added and each condition contains $V_{\mathrm{c}}(V_{\mathrm{c}}-1)$ training sequences.
Copying predicts the visible final content token, induction predicts the token following a repeated anchor, and random-label memorization assigns each unique sequence one fixed, uniformly sampled content-token label.

\paragraph{Training grid.}
The systematic grid crosses three tasks, depths $\{1,2,3\}$, lengths $\{4,6,8,10,12\}$, and content vocabularies $\{6,8,12,16\}$, subject to $V_{\mathrm{c}}\ge L-1$, with five seeds per valid condition (675 runs).
The seed determines sampled sequences, fillers, anchor positions, or random labels as applicable.
Models train for 100k steps using full-batch SGD at learning rate $0.01$ and last-position cross-entropy, with parameters recorded every 100 steps.
No train--test split is used.

\paragraph{Behavioral concentration and reconstruction (in-sample).}
PR and reconstruction are evaluated in-sample using all $V_{\mathrm{c}}+4$ last-position logits on the finite training sequences.
The complete recorded trajectory defines the SVD basis.
Behavioral-energy statistics use the outgoing intervals at 1k, 5k, 10k, 20k, and 50k, with JVPs estimated by forward finite differences at $\epsilon=0.001$.
At each checkpoint, PR uses the full realized-contribution energy distribution.
Trajectory-level reconstruction ordering aggregates contributions across the five intervals.
Top-$K$ reconstruction uses this trajectory-level behavioral-energy ordering and is evaluated at 50k in the reference configuration (depth 1, length 4, content vocabulary 6).

\subsection{Training and Behavioral Concentration Diagnostics}

\begin{figure}[h]
\centering
\includegraphics[width=0.92\linewidth]{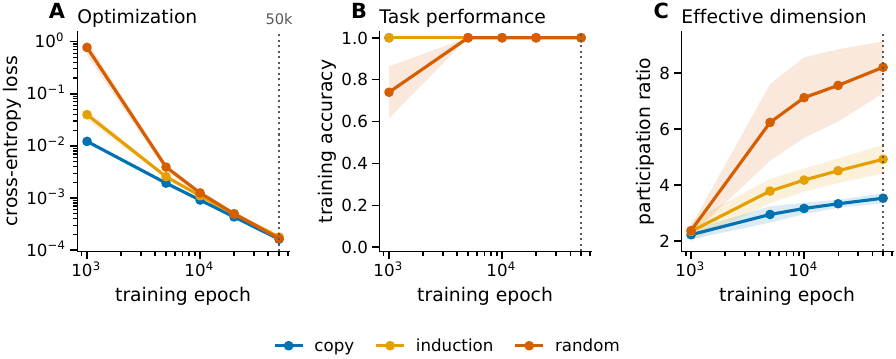}
\caption{\textbf{Next-token optimization and participation-ratio trajectories.}
\textbf{A--C:} Cross-entropy, training accuracy, and participation ratio over time for copying, induction, and random labels in the matched reference configuration (depth 1, sequence length 4, content vocabulary 6).
Lines and shading are mean $\pm$ standard deviation over five seeds, and the dotted line marks the 50k checkpoint used for the main-text task/data comparison.}
\label{fig:supp:transformer-training}
\end{figure}

Figure~\ref{fig:supp:transformer-training} shows that copying reaches a compact modal distribution early, whereas induction and random-label memorization recruit additional modes as training proceeds.
All tasks reach perfect training accuracy by the endpoint, so the PR differences describe how learning is distributed rather than differences in final fit.

\begin{figure}[h]
\centering
\includegraphics[width=0.90\linewidth]{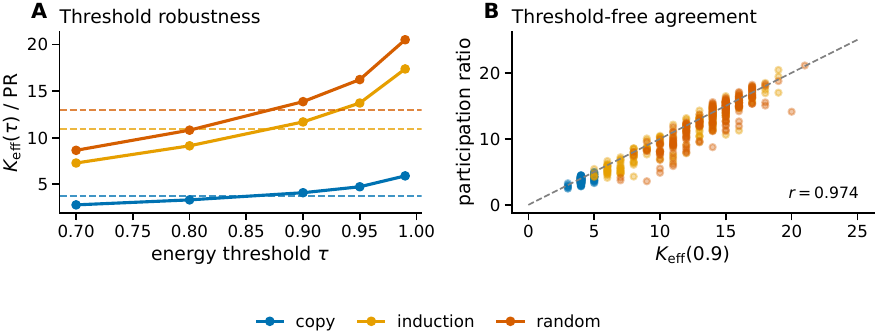}
\caption{\textbf{Next-token behavioral-concentration diagnostics.}
\textbf{A:} $K_{\mathrm{eff}}(\tau)$ varies with the selected energy threshold from 0.70 to 0.99, with dashed horizontal lines showing the threshold-independent mean PR for each task.
\textbf{B:} PR strongly agrees with the auxiliary $K_{\mathrm{eff}}(0.9)$ count.}
\label{fig:supp:transformer-diagnostics}
\end{figure}

Figure~\ref{fig:supp:transformer-diagnostics} shows that task ordering is stable across cumulative-energy thresholds and that PR tracks $K_{\mathrm{eff}}(0.9)$ while remaining threshold-free.

\begin{figure}[h]
\centering
\includegraphics[width=0.58\linewidth]{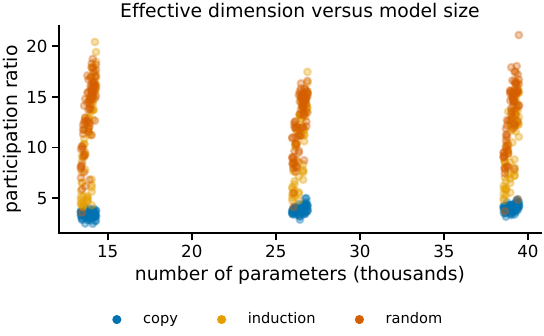}
\caption{\textbf{Behavioral PR versus parameter count.}
Across the controlled grid (approximately 13k--39k parameters), PR separates more strongly across task groups than across the displayed parameter-count range.
This is a descriptive bounded-grid association rather than a variance decomposition or claim of scale invariance.}
\label{fig:supp:parameter-count}
\end{figure}

Figure~\ref{fig:supp:parameter-count} compares behavioral PR with parameter count over the bounded grid.
Restricted-support checks preserve the task ordering despite the coupling between sequence length $L$ and content vocabulary $V_{\mathrm{c}}$.
With $V_{\mathrm{c}}\in\{12,16\}$ ($n=150$ per task), mean PR is $3.8\pm0.5$, $12.5\pm3.5$, and $14.7\pm2.2$ for copying, induction, and random labels, respectively.
With $L\in\{4,6\}$ ($n=120$ per task), the corresponding values are $3.7\pm0.4$, $8.3\pm2.9$, and $12.0\pm3.4$.
Within these marginal summaries, mean PR varies less across depth levels than across task constructions.
The full grid contains 15 valid length--vocabulary pairs satisfying $V_{\mathrm{c}}\ge L-1$.
Consequently, the main figure's marginal length and vocabulary curves are associated summaries rather than independent interventions.

\subsection{Trajectory dimension and static endpoint rank}
\label{app:dynamic-static}

Trajectory concentration and endpoint low rank are related but distinct objects.
Mode Dynamics measures how many trajectory-derived directions carry behavioral change during training, whereas low-rank adaptation (LoRA) imposes a static low-rank parameterization \citep{hu2021lora}.
We compare behavioral PR at epoch 50k with a separate endpoint head-spectrum PR computed from the singular-value energy of the output-head change from initialization.

\begin{figure}[h]
\centering
\includegraphics[width=0.82\linewidth]{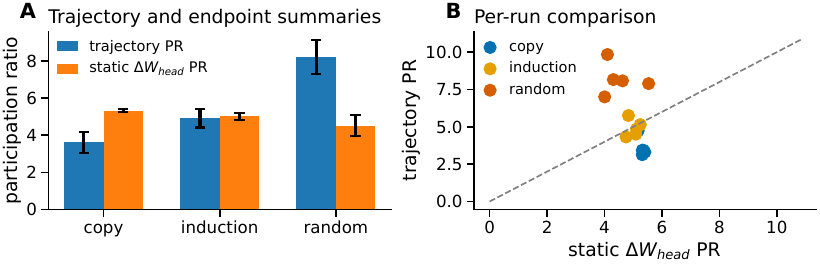}
\caption{\textbf{Dynamic trajectory and static endpoint summaries.}
\textbf{A:} Behavioral PR of realized-contribution energy at epoch 50k and endpoint head-spectrum PR across tasks.
\textbf{B:} Per-run comparison.
The quantities are not commensurate.
The dynamic statistic uses full-parameter realized contributions at 50k, whereas the static statistic uses the 0--100k output-head change and is capped by head rank.}
\label{fig:supp:dynamic-static}
\end{figure}

Figure~\ref{fig:supp:dynamic-static} shows similar static output-head dimensionality across tasks despite descriptive differences in trajectory-level behavioral PR, so endpoint head rank does not substitute for the trajectory analysis and is not a quantitative rank-gap estimate.
This diagnostic does not evaluate LoRA fine-tuning, mode rotation, or cross-seed alignment.

\FloatBarrier
\subsection{Per-Example Top-\texorpdfstring{$K$}{K} Behavioral Reconstruction}
\label{app:heatmaps}

\begin{figure}[h]
\centering
\includegraphics[width=0.84\linewidth,trim=0 0 0 126.4,clip]{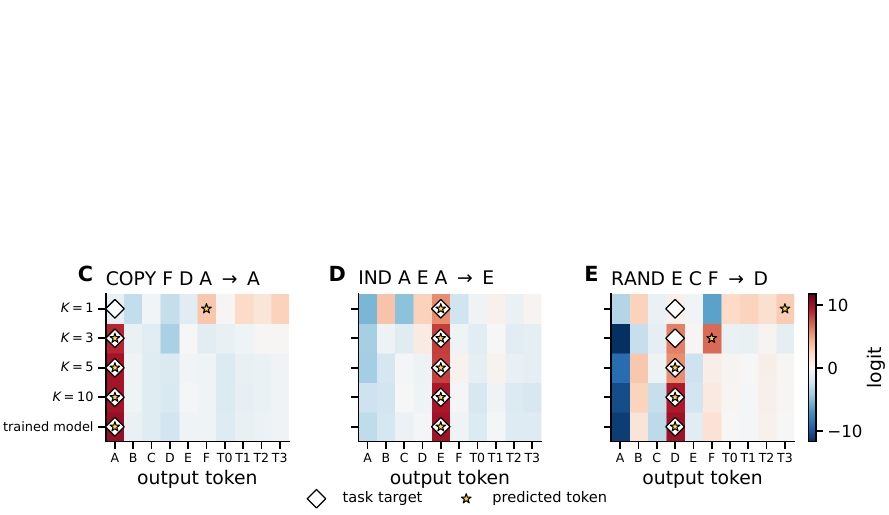}
\caption{\rev{\textbf{Highest-ranked per-example cumulative logit reconstructions.}}
\textbf{C--E:} Selected training examples illustrate low-to-high-$K$ recovery at epoch 50k.
Rows progress through cumulative Top-$K$ modal reconstructions for $K\in\{1,3,5,10\}$ to the same trained full-network output.
Diamonds mark targets and stars mark row predictions, overlapping when correct.}
\label{fig:supp:reconstruction-examples}
\end{figure}

\begin{figure}[h]
\centering
\includegraphics[width=0.92\linewidth]{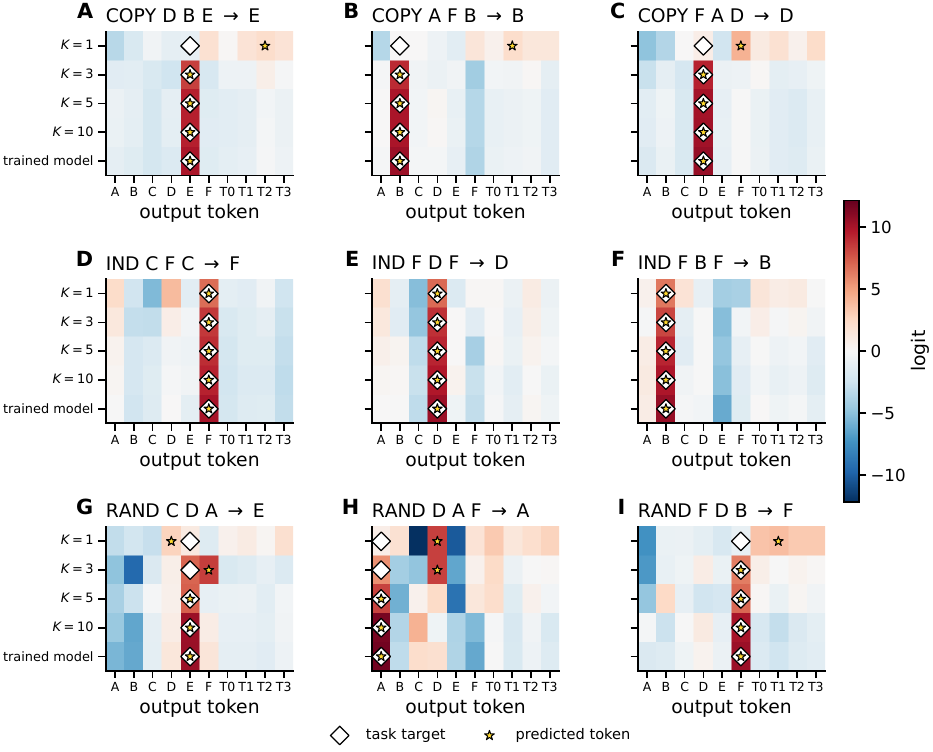}
\caption{\textbf{Selected per-example cumulative logit reconstructions.}
\textbf{A--C:} Copying.
\textbf{D--F:} Induction.
\textbf{G--I:} Random labels.
Each subplot is one training example.
Rows progress through cumulative Top-$K$ modal reconstructions to the trained-model full-network reference.
White diamonds and smaller gold stars are centered on the task target and row prediction, respectively, so a correct prediction appears as a star over a diamond.}
\label{fig:supp:heatmaps}
\end{figure}

Among examples with correct trained-model and Top-10 predictions and monotonically improving per-example logit $R^2$, we rank by the number of prediction changes, then the number of distinct predictions, and then the $K=1$-to-Top-10 improvement.
\rev{Figure~\ref{fig:supp:reconstruction-examples} shows the highest-ranked case.}
Figure~\ref{fig:supp:heatmaps} shows three additional cases per task in a $3\times3$ layout, while preferring distinct target tokens.
Induction has no prediction changes, consistent with its perfect aggregate $K=1$ agreement, but its selected examples still show improving logits.
These behavioral reconstructions do not test circuit localization or semantic monosemanticity.

\FloatBarrier
\rev{Across these selected examples, copying and induction often recover the correct decision before every logit is numerically reconstructed, whereas random-label examples generally require broader sets of leading modes.}

\FloatBarrier
\begin{revblock}
\subsection{Causal Suppression of Leading Modes}
\label{app:causal}

\paragraph{Task and split.}
The causal test uses induction with sequence length $L=6$ and content vocabulary $V_{\mathrm{c}}=6$.
Each context is a task token followed by five content tokens, in which an anchor and its target appear consecutively, two distinct fillers occupy the remaining interior positions, and the anchor repeats at the final position.
The 30 anchor--target pairs, three anchor positions, and 12 ordered filler pairs give 1,080 valid contexts.
For each of five seeds, a seeded random split assigns 864 contexts to training and mode ranking and 216 to testing.
Every anchor--target pair appears in both splits, so held-out contexts differ in anchor position and fillers, and they are never used for training or ranking.

\paragraph{Reference run and ranking.}
The model is the depth-1 transformer of Appendix~\ref{app:transformers} with 13,472 parameters, trained for 10,000 steps by deterministic full-batch SGD at learning rate $0.01$, with parameters recorded every 10 steps.
The 1,000 recorded updates of this reference run define the SVD basis.
Modes follow behavioral-energy ordering (Equation~\ref{eq:trajectory-energy}) over the outgoing intervals at steps 100, 500, 1k, 2k, and 5k, using all last-position logits on the training contexts and forward finite-difference JVPs with $\epsilon=0.001$.

\paragraph{Suppression and controls.}
Each suppressed run restarts from the reference run's initial parameters and training contexts and replaces every gradient $\nabla_t$ by $\nabla_t-U_KU_K^\top\nabla_t$, where the $K$ orthonormal columns of $U_K$ span the suppressed directions.
Top-$K$ suppression uses the $K$ leading behavioral modes, $K\in\{1,3,5\}$.
For the energy-matched control, each of the five leading modes is matched in turn to the unused mode outside the Top-5 whose parameter-motion energy $\sigma_k^2$ is closest on a logarithmic scale, and the $K$-mode control uses the first $K$ matches.
The tensor-matched control applies one random permutation and sign flip of coordinates within each parameter tensor, shared by all modes, to the Top-$K$ modes.
This preserves the subspace dimension, pairwise inner products, and each mode's norm within every tensor, but changes which parameters move together.
An unsuppressed replay reproduced every reference loss and recorded parameter vector exactly, and the relative gradient component remaining in a suppressed subspace stayed below $3\times10^{-7}$.

\paragraph{Results.}
Table~\ref{tab:supp:causal} reports held-out accuracy at step 7,000, by which every baseline run had crossed $90\%$, and at the final step 10,000.
In every seed and for every $K$, Top-$K$ suppression ended below both matched controls at step 7,000.
The impairment is not monotonic in $K$, so these data support the importance of these particular modes rather than a dose response.
Training stopped at step 10,000, so these runs do not show whether suppressed models would eventually recover.

\begin{table}[h]
\color{\revcolor}
\centering
\caption{Held-out induction accuracy (\%) under causal suppression, as mean $\pm$ population standard deviation over five seeds.
The baseline does not depend on $K$.}
\label{tab:supp:causal}
\small
\setlength{\tabcolsep}{4pt}
\begin{tabular}{lcccccc}
\toprule
& \multicolumn{3}{c}{Step 7,000} & \multicolumn{3}{c}{Step 10,000} \\
\cmidrule(lr){2-4}\cmidrule(lr){5-7}
Suppressed modes & $K=1$ & $K=3$ & $K=5$ & $K=1$ & $K=3$ & $K=5$ \\
\midrule
None (baseline) & \multicolumn{3}{c}{$95.7\pm3.6$} & \multicolumn{3}{c}{$99.4\pm1.1$} \\
Top-$K$ & $29.8\pm13.3$ & $21.5\pm14.0$ & $38.6\pm5.5$ & $34.6\pm17.9$ & $35.6\pm22.0$ & $48.3\pm12.0$ \\
Energy-matched & $89.1\pm17.8$ & $86.1\pm18.2$ & $87.0\pm20.3$ & $93.2\pm11.0$ & $93.7\pm11.0$ & $98.0\pm2.4$ \\
Tensor-matched & $95.7\pm3.6$ & $96.0\pm3.3$ & $95.8\pm3.4$ & $99.5\pm0.7$ & $99.5\pm0.7$ & $99.5\pm0.7$ \\
\bottomrule
\end{tabular}
\end{table}
\end{revblock}

\begin{revblock}
\subsection{Initialization and Checkpoint-Density Robustness}
\label{app:robustness}

\paragraph{Setting and metrics.}
Both analyses use the depth-1 induction condition with $L=V_{\mathrm{c}}=6$ and its 30 training sequences, one per anchor--target pair, under the systematic-grid protocol above.
Each run trains for 100k full-batch SGD steps at learning rate $0.01$ and records parameters every 100 steps, giving 1,000 updates.
Modes follow behavioral-energy ordering over the outgoing intervals at 1k, 5k, 10k, 20k, and 50k on the training sequences, so all metrics are in-sample.
PR and $K_{\mathrm{eff}}(0.9)$ summarize the trajectory-level shares $\bar\rho_k$ of Equation~\eqref{eq:trajectory-energy} rather than the checkpoint-specific shares behind Figure~\ref{fig:transformer-dynamics}.
Top-5 $R^2$ applies Equation~\eqref{eq:r2} to vocabulary-centered net contributions, with $s=\sum_k G_k$ summed over all modes and $\widehat s$ the same sum over the five leading modes.

\paragraph{Initialization.}
Five runs share the training sequences, example order, optimizer, and checkpoint schedule and differ only in the initialization seed.
Table~\ref{tab:supp:robustness} shows that trajectory-level concentration stays compact in every run.
The numerical rank of each update matrix, the number of singular values above $T\epsilon\,\sigma_1$ with $T=1{,}000$ and float32 machine precision $\epsilon$, ranges from 47 to 52.
Because each run has its own basis, this analysis tests whether each run is compact, not whether individual modes align across initializations or activate in the same order.

\paragraph{Checkpoint density.}
Five reference runs with seeds 0--4, which vary both the generated sequences and the initialization, are subsampled to nested schedules that keep every first, second, fifth, or tenth recorded checkpoint.
These schedules have spacings of 100, 200, 500, and 1,000 steps, giving 1,000, 500, 200, and 100 updates.
Each schedule recomputes its SVD basis and behavioral ordering from the outgoing intervals at the same five ranking checkpoints, so these intervals lengthen as spacing grows.
Over this tenfold refinement, mean $K_{\mathrm{eff}}(0.9)$ rises from 2.0 to 3.0 and mean PR from 1.66 to 1.97, while Top-5 $R^2$ remains at least 0.968 in every run (Table~\ref{tab:supp:robustness}).
The leading subspaces are less stable.
Relative to 100-step spacing, the mean overlap of the five leading modes, measured as the average squared cosine of the principal angles between the two subspaces, is 0.850, 0.743, and 0.527 at 200-, 500-, and 1,000-step spacing.
The same overlap between the spans of their net contributions $G_k$ is 0.830, 0.595, and 0.431.
Individual modes therefore depend on the schedule, while the small count nearly stabilizes.
Every-step analysis would require $10^5$ updates per run and was not performed.

\begin{table}[h]
\color{\revcolor}
\centering
\caption{Trajectory-level concentration in the controlled induction condition ($L=V_{\mathrm{c}}=6$, 30 training sequences), as mean $\pm$ population standard deviation over five runs.
Brackets give the range of $K_{\mathrm{eff}}(0.9)$ and the minimum Top-5 $R^2$.}
\label{tab:supp:robustness}
\small
\setlength{\tabcolsep}{4pt}
\begin{tabular}{lrrlll}
\toprule
Varied factor & Spacing & Updates & PR & $K_{\mathrm{eff}}(0.9)$ & Top-5 $R^2$ \\
\midrule
Initialization only & 100 & 1,000 & $1.99\pm0.52$ & $3.6\pm1.2$ [2--5] & $0.974\pm0.015$ [0.949] \\
\addlinespace[2pt]
Checkpoint spacing & 1,000 & 100 & $1.66\pm0.17$ & $2.0\pm0.0$ [2--2] & $>0.9999$ \\
 & 500 & 200 & $1.68\pm0.15$ & $2.0\pm0.0$ [2--2] & $0.998\pm0.001$ [0.995] \\
 & 200 & 500 & $1.72\pm0.20$ & $2.2\pm0.4$ [2--3] & $0.989\pm0.006$ [0.980] \\
 & 100 & 1,000 & $1.97\pm0.32$ & $3.0\pm0.6$ [2--4] & $0.979\pm0.009$ [0.968] \\
\bottomrule
\end{tabular}
\end{table}
\end{revblock}

\FloatBarrier
\section{Pixel-Space Reconstruction on CIFAR-100}
\label{app:cifar}

\subsection{Protocol and Reproducibility}

\paragraph{Task and model.}
This experiment analyzes pixel-level reconstruction rather than classification.
We train a convolutional autoencoder on all 50,000 CIFAR-100 training images using mean absolute RGB pixel loss.
Class logits receive zero loss weight.
Inputs use no augmentation and are normalized by channel means $(0.5071,0.4867,0.4408)$ and standard deviations $(0.2675,0.2565,0.2761)$.
The encoder applies four convolution--group-normalization--ReLU stages with 48, 96, 192, and 192 channels and one max-pool downsampling to a $16\times16$ spatial bottleneck.
The no-skip decoder bilinearly upsamples once, applies a convolution--group-normalization--ReLU stage, and uses a sigmoid RGB head.
Neither side has residual or refinement blocks.
The model has 730,087 trainable parameters.

\paragraph{Training and design selection.}
All five runs use Adam with learning rate $10^{-3}$, $(\beta_1,\beta_2)=(0.9,0.999)$, numerical-stability constant $10^{-8}$, no weight decay, batch size 256, and 50 epochs.
Training uses shuffled minibatches on Apple MPS, with NumPy and PyTorch seeds 0--4 controlling initialization and sampling.
The architecture, optimizer, integration settings, and $K=50$ endpoint were selected using seed-0 decoder fidelity (including endpoint test-image L1), training-split integration calibration, and exploratory modal analyses before running seeds 1--4.
Seed 0 remains in the five-run aggregate, so the uncertainty summary is descriptive and includes the design seed.
No test-image modal reconstruction score was inspected before the protocol, source hashes, and exact test indices were frozen.

\paragraph{Evaluation.}
The CIFAR development analysis evaluates modal fidelity in-sample on a deterministic, seed-specific class-balanced subset of 100 training images, one per CIFAR-100 class.
The locked evaluation instead evaluates the already selected Top-50 reconstruction on a fixed class-balanced subset of 100 test images per seed.
These images are held out from training, modal scoring, mode selection, and reranking, but not from the earlier endpoint decoder-L1 check.
A separate eight-image seed-0 training-split analysis supplies the full $K\in\{1,3,5,10,20,30,40,50\}$ curve and qualitative examples.
The class-balanced training and test subsets use the stated training seed plus 10,000 and 20,000, respectively.
Randomized SVD uses the training seed.

\subsection{Path-Integrated Reconstruction and Metrics}

\paragraph{Trajectory basis and retained parameter motion.}
There are 196 optimizer steps per epoch.
Storing parameters every five steps gives 1,961 checkpoints and $N=1{,}960$ update intervals.
For $i=0,\ldots,N-1$, let $\Delta\theta_i=\theta_{i+1}-\theta_i$.
The main framework stacks these updates as columns of $V$.
The implementation forms $D=V^\top\in\mathbb{R}^{N\times P}$, where $P=730{,}087$, and estimates its first 50 right singular vectors using scikit-learn randomized SVD with four power iterations and \texttt{random\_state=seed}.
These right singular vectors are the parameter modes $u_k$ defined from $V$, and $\Pi_K=\sum_{k=1}^{K}u_ku_k^\top$ projects onto the first $K$ modes.
By Equations~\eqref{eq:amplitude}--\eqref{eq:decomposition}, the parameter-motion energy $\sigma_k^2=\sum_i a_{k,i}^2$ is the total squared motion along mode $k$.
Figure~\ref{fig:cifar-rgb}B reports the retained fraction $\sum_{k\le K}\sigma_k^2/\lVert V\rVert_F^2$, whose denominator includes all recorded updates rather than only the 50 estimated modes.
This parameter-space diagnostic is distinct from the behavioral cumulative share $C_K(t)$ defined in Section~\ref{sec:concentration}.

\paragraph{Path-integrated output reconstruction.}
The Top-$K$ RGB-change reconstruction integrates the projected updates along the straight checkpoint chords:
\begin{equation}
\begin{aligned}
\widehat{\Delta f}_K(x)
&=\sum_{i=0}^{N-1}\int_0^1
J_\theta f\!\left(x,\theta_i+c\Delta\theta_i\right)
\Pi_K\Delta\theta_i\,dc \\
&\approx
\sum_{i=0}^{N-1}\sum_{q=1}^{Q}w_q
J_\theta f\!\left(x,\theta_i+c_q\Delta\theta_i\right)
\Pi_K\Delta\theta_i .
\end{aligned}
\label{eq:cifar-path-integral}
\end{equation}
Here $Q=4$ and $(c_q,w_q)$ are the Gauss--Legendre nodes and weights on $[0,1]$.
Because $\Pi_K\Delta\theta_i=\sum_{k\le K}a_{k,i}u_k$, the integrand sums the main-text realized contributions $a_{k,i}\phi_k(x,\theta_i+c\Delta\theta_i)$ along the chord.
Thus Equation~\eqref{eq:cifar-path-integral} is the path-integrated counterpart of the main modal reconstruction, not a new decomposition.
JVPs are computed directly with forward-mode AD and batched over the selected images.
On eight seed-0 images, full-update calibration over two 100-step windows gave chord-integration $R^2=0.983$ for steps 0--100 and $0.999$ for steps 96--196 at the chosen five-step spacing and four-point rule.
This checks short-window numerical integration, not a full-basis reconstruction ceiling over all 50 epochs.

\paragraph{Fidelity metrics.}
The reconstruction reference is the trained decoder's output change $\Delta f(x)=f(x,\theta_{\rm final})-f(x,\theta_0)$, and adding $\widehat{\Delta f}_K(x)$ to the initial output gives the reconstructed final decoder output.
We apply Equation~\eqref{eq:r2} with $s=\Delta f$ and $\widehat s=\widehat{\Delta f}_K$, stacked over the selected images, pixels, and RGB channels.
Thus the reference is the trained decoder's RGB change, not the ground-truth image or class label.
Per-image $R^2$ applies the same equation separately to each image.
Because its denominator can be small for nearly constant image-wise changes, we additionally record $\|\widehat{\Delta f}_K-\Delta f\|_2/\|\Delta f\|_2$ and emphasize the pooled score and distributional quantiles rather than the minimum.
The implementation adds $10^{-12}$ to variance and norm denominators for numerical stability.
Decoder mean absolute RGB error instead compares $f(x,\theta_{\rm final})\in[0,1]$ with the ground-truth image and answers a separate question.

\subsection{Signed-Permutation (Norm-Matched) Coordinate Control}

Each control draw samples one parameter permutation and one independent Rademacher sign vector, initialized with the training seed plus 30,000.
The same fixed signed permutation is applied to the Top-50 projected update at every interval and quadrature node.
This transformation exactly preserves each projected update's $\ell_2$ norm and all temporal inner products while randomizing its global parameter-coordinate assignment.
Three independent draws are evaluated on seed 0.
The controls therefore test whether learned coordinate alignment matters at matched instantaneous update norm.
They do not provide five-seed uncertainty for the random baseline.
They were fixed and scored in the training-split development analysis and were not recomputed on the held-out test images.
Because the transformation crosses parameter tensors and can move energy into the RGB-irrelevant classifier head, it is neither layer- nor support-matched and does not isolate fine-grained trajectory specificity.
\rev{Appendix~\ref{app:cifar-tensor} adds a held-out control that is both layer- and support-matched.}

\begin{revblock}
\subsection{Tensor-Matched Coordinate Control}
\label{app:cifar-tensor}

This post-hoc control was run after the locked held-out evaluation and leaves it unchanged.
For each of the five seeds, we deterministically replayed the locked training trajectory and required its 50 leading singular values to match the frozen spectrum and its same-run Top-50 held-out $R^2$ to match the locked score within $10^{-4}$.
Each of three draws per seed samples one random permutation and one independent Rademacher sign vector inside every parameter tensor that affects the RGB output, initialized with the training seed plus 30,000.
Classifier-head coordinates are left unchanged, and a JVP check confirmed that they do not affect the RGB output.
The same transform is applied to the Top-50 projected update at every interval and quadrature node, which is equivalent to transforming all 50 modes with one shared transform.
It therefore preserves each mode's norm within every tensor and all inner products between modes, moves no motion across tensors or into the head, and changes only which parameters within each tensor move together.
Without retraining, the transformed updates are path-integrated with the same four-node rule on the same 100 frozen held-out test images per seed.
Across the 15 seed--draw pairs, pooled RGB-change $R^2$ is $-0.909\pm0.559$, with every draw below zero, compared with $0.912\pm0.012$ for the Top-50 reconstruction over the five seeds.
Every paired difference exceeds 0.9.
Recovering the RGB change therefore depends on which parameters move together within each tensor, not only on how much each tensor moves.
We did not run the corresponding Pythia control.
\end{revblock}

\subsection{Additional Results}

\begin{figure}[h]
\centering
\includegraphics[width=0.92\linewidth]{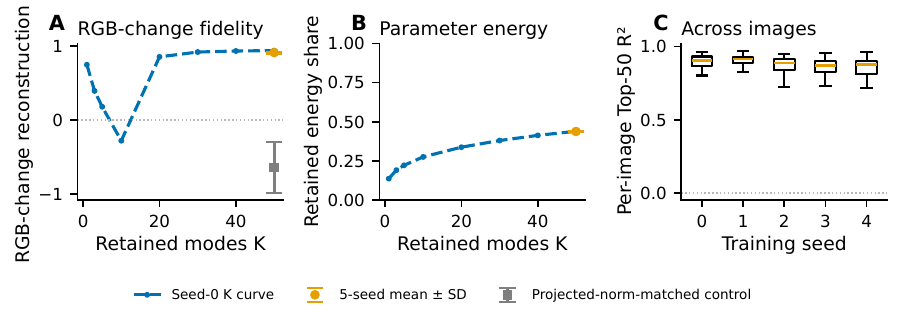}
\caption{\textbf{A small parameter-energy-ranked subspace reconstructs the RGB change induced by CIFAR-100 training.}
\textbf{A:} Training-split seed-0 eight-image $K$ curve, five-seed Top-50 mean $\pm$ population standard deviation (100 training images per seed), and three seed-0 norm-matched controls.
\textbf{B:} Fraction of parameter-motion energy retained.
\textbf{C:} Training-split per-image Top-50 $R^2$ (100 images per box), with extreme fliers hidden only visually.
Panel A reconstructs the trained decoder's RGB change from initialization, not the CIFAR target, and modal scoring uses training images.}
\label{fig:cifar-rgb}
\end{figure}

\paragraph{Aggregate fidelity.}
The five training-split pooled Top-50 scores underlying Figure~\ref{fig:cifar-rgb} are $0.922$, $0.934$, $0.911$, $0.908$, and $0.892$.
Excluding the design seed gives $0.911\pm0.015$ over seeds 1--4, leaving the descriptive conclusion unchanged.
The locked held-out scores are $0.924$, $0.927$, $0.902$, $0.896$, and $0.908$, giving $0.912\pm0.012$ across seeds while retaining $43.8\%\pm0.4\%$ of parameter-motion energy.
Pooling the 500 held-out seed--image evaluations gives median per-image $R^2=0.892$, interquartile range $[0.839,0.919]$, and tenth percentile $0.763$.
Overall, $99.4\%$ of scores are positive.
The held-out median per-image relative $\ell_2$ error is $0.291$ with interquartile range $[0.250,0.335]$.
The Panel A control spread is over three seed-0 random draws, whereas the Top-50 error bar is over five training seeds.

\begin{figure}[h]
\centering
\includegraphics[width=0.92\linewidth]{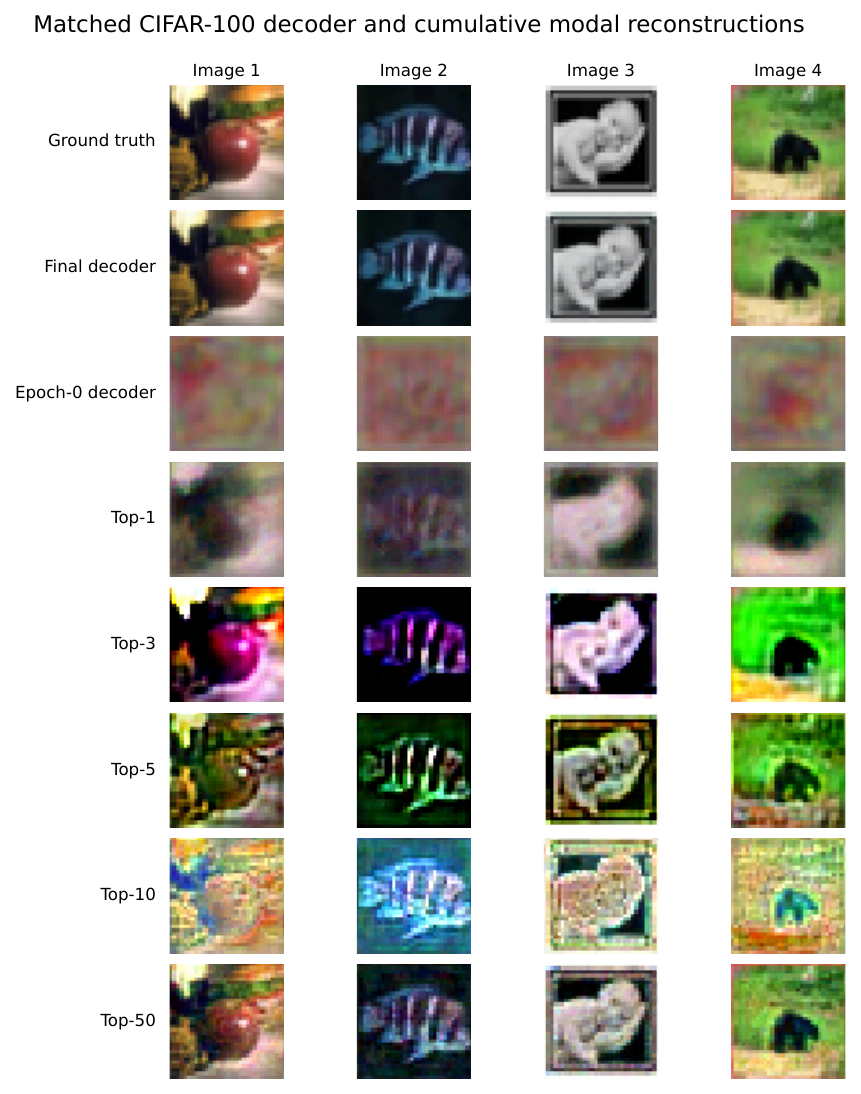}
\caption{\textbf{Matched CIFAR-100 decoder outputs and cumulative modal reconstructions.}
Columns are fixed evaluation images.
Rows show the ground truth, final decoder, epoch-0 decoder, and cumulative Top-1, Top-3, Top-5, Top-10, and Top-50 path-integrated predictions from the detailed seed-0 run.
These examples illustrate pixel-level recovery, while aggregate claims use the five-seed, 100-image results in Figure~\ref{fig:cifar-rgb}.}
\label{fig:supp:cifar-images}
\end{figure}

\paragraph{Exploratory \texorpdfstring{$K$}{K}-dependence and examples.}
The detailed seed-0 curve in Figure~\ref{fig:cifar-rgb}A is non-monotonic at small and intermediate $K$, including negative $R^2$ at $K=10$.
Parameter-energy ordering does not guarantee monotonic output-space reconstruction, so the curve documents the observed progression rather than implying that every additional mode improves fidelity.
Figure~\ref{fig:supp:cifar-images} shows the corresponding cumulative reconstructions for fixed examples.

\FloatBarrier
\section{Public Pythia-70M Pretraining Trajectory}
\label{app:pythia}

\subsection{Protocol and Data Separation}

\paragraph{Model and checkpoints.}
Pythia-70M is a decoder-only GPT-NeoX-style transformer released with intermediate pretraining checkpoints \citep{biderman2023pythia}.
We analyze those released checkpoints without additional training or fine-tuning.
Models and interpolated parameters are evaluated in float32 with eager attention on Apple MPS.
The analysis uses 51 fixed checkpoints spanning step 0 through step 143000, with dense sampling through step 512 and broad coverage thereafter.
The exact early schedule is $0,4,8,16,32,64,128,256,512$, followed by 1000, 2000, and $1000\,\mathrm{round}(1+142i/40)$ for $i=1,\ldots,40$.
Thus the later intervals span either 3,000 or 4,000 optimizer steps.
Flattening every trainable tensor in a fixed manifest gives 70,426,624 parameters per checkpoint and 50 checkpoint-to-checkpoint updates.
Eigendecomposition of the $50\times50$ row Gram matrix yields a numerically full-rank 50-mode basis.
The maximum observed orthonormality error is $1.08\times10^{-5}$.

\paragraph{Evaluation contexts.}
We tokenize without adding special tokens and deterministically select 256 nonoverlapping blocks of 129 tokens from the official Pile validation split \citep{gao2020pile}.
A SHA-256 hash of the fixed seed, source line number, and text determines candidate-document order and each token offset.
The validation and test procedures scan 1,024 and 512 candidates, respectively, accepting the first nonoverlapping blocks.
For each block, the first 128 tokens form the input prefix and the final token is retained only for task-level diagnostics.
The first 128 blocks define the behavioral ranking set, and the remaining 128 form a disjoint validation scoring set used for development.
Before any Pile-test access, we froze the validation-derived ordering, deterministic test-context selector, retained mode budgets, controls, two-node quadrature rule, and primary pooled $R^2$ metric.
The confirmatory evaluation then deterministically selects 128 blocks from the previously untouched official Pile test split.
The test procedure creates no ranking set and performs no reranking or tuning.
Stored line numbers, token offsets, selection hashes, source-file digests, and frozen artifact hashes provide an audit trail.

\subsection{Behavioral Ordering and Path Reconstruction}

\paragraph{Ordering.}
The trajectory modes are defined only by the parameter trajectory.
To obtain the behavioral-energy ordering, we integrate each mode's realized contribution to the 50,304 final-position logits over all 50 checkpoint chords on the ranking contexts.
Each chord uses two-point Gauss--Legendre quadrature and forward-mode JVPs.
The two nodes are $c=(1\pm1/\sqrt{3})/2$ on the unit chord, each with weight $1/2$.
At each node $c$, we evaluate the model at interpolated parameters $\theta_i+c\Delta\theta_i$.
These are analysis points, not recovered intermediate training checkpoints.
We subtract each effect's mean across vocabulary before applying Equation~\eqref{eq:trajectory-energy}, removing the additive logit direction that leaves softmax probabilities unchanged.
The ranking therefore uses full output-vector effects but no true next-token labels, losses, or decisions.
Once estimated, the order is frozen for all scoring analyses.

\paragraph{Held-out reconstruction and controls.}
On the untouched Pile-test contexts, we path-integrate cumulative validation-ranked subsets with $K\in\{1,3,5,10,20,50\}$ and compare them with the actual initialization-to-final centered-logit change.
Let $C(z)=z-\frac{1}{V}\sum_j z_j$ center one context's vocabulary logits.
The pooled score is
\begin{equation}
R^2_{\mathrm{pooled}}
=
1-
\frac{\sum_c\lVert C(\widehat{\Delta z}_c)-C(\Delta z_c)\rVert_2^2}
{\sum_c\lVert C(\Delta z_c)\rVert_2^2}.
\label{eq:pythia-pooled-r2}
\end{equation}
Its 95\% interval resamples the 128 contexts with replacement 2,000 times using seed 10,070 and recomputes this ratio while retaining all vocabulary coordinates within each sampled context.
The full-update reconstruction integrates complete checkpoint updates along the path and supplies a numerical reference without modal projection.
The initialization-frozen Top-$K$ curve instead evaluates each cumulative $K$-mode displacement at the initial parameters.
Its $K=50$ endpoint is the initialization-frozen all-mode reconstruction up to numerical precision and remains distinct from full-update.
A fixed signed parameter permutation of the 50-mode projected update preserves its norm at every interval but disrupts learned coordinate alignment.

\subsection{Results and Boundaries}

\begin{figure}[h]
\centering
\includegraphics[width=0.92\linewidth]{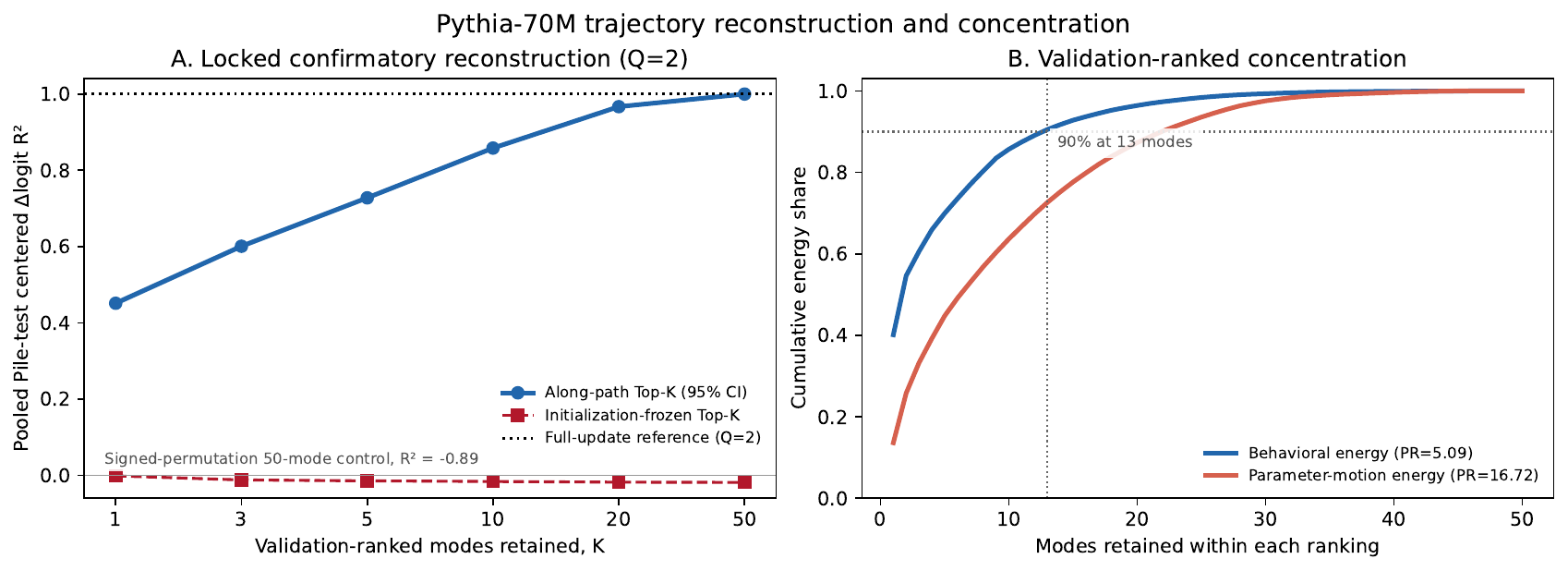}
\caption{\textbf{Pythia-70M centered-logit change is concentrated in a behaviorally ranked trajectory subspace.}
\textbf{A:} Locked confirmatory Pile-test pooled centered-$\Delta$logit $R^2$ across validation-ranked $K$-mode subsets.
Blue evaluates effects along the checkpoint path.
Red is the initialization-frozen Top-$K$ reconstruction, with $K=50$ equal to the initialization-frozen all-mode reconstruction.
The dotted line is full-update.
Shading is a 95\% context-bootstrap interval, and the annotation gives the signed-permutation 50-mode control.
\textbf{B:} Validation-ranking cumulative behavioral energy in behavioral-energy ordering and cumulative parameter-motion energy in parameter-energy ordering.
Legends report PR.
Guides mark the 90\% behavioral-energy threshold at 13 modes.
The curves use separate rankings and are not a common-order comparison.}
\label{fig:supp:pythia-reconstruction}
\end{figure}

Figure~\ref{fig:supp:pythia-reconstruction}A shows untouched-test pooled centered-$\Delta$logit $R^2$ values of $0.451$, $0.601$, $0.728$, $0.858$, $0.9668$, and $0.9998$ for $K=1,3,5,10,20,50$, respectively.
The $K=20$ 95\% context-bootstrap interval is $[0.9662,0.9673]$.
The full-update integration reference is $0.999809$.
The all-mode ($K=50$) pooled score differs from the full-update reference by less than $10^{-7}$.
The signed-permutation 50-mode control gives $-0.890$, and the initialization-frozen all-mode reconstruction gives $-0.018$.

On the frozen validation ranking set, behavioral energy is more concentrated than parameter motion.
Behavioral PR is $5.09$ with $K_{\mathrm{eff}}(0.9)=13$, compared with PR $16.72$ for the parameter-motion energy shares $\sigma_k^2/\sum_j\sigma_j^2$.
The behavioral-energy and parameter-energy rank positions have Spearman correlation $0.878$, indicating substantial overlap without equivalence.

\begin{revblock}
\paragraph{Post-hoc same-budget ordering comparison.}
To test whether this difference in ordering matters for held-out reconstruction, we rescored the same 128 frozen Pile-test contexts after retaining the same numbers of modes in parameter-energy ordering, using the locked 51-checkpoint basis and two-node quadrature.
This comparison was run after the locked result and does not change it, and it performs no ranking on test data.
Table~\ref{tab:pythia-ordering} shows that behavioral-energy ordering reconstructs centered logit change much better at small budgets, with $R^2=0.728$ versus $0.134$ for five modes, and that the gap narrows as more modes are retained.
It does not uniformly improve distributions or decisions.
Parameter-energy ordering gives lower KL divergence up to $K=10$ and slightly higher Top-1 prediction agreement for $K\le5$, whereas behavioral-energy ordering gives lower KL divergence at $K=20$.
The two orderings coincide at $K=50$.

\begin{table}[h]
\color{\revcolor}
\centering
\caption{\textbf{Post-hoc same-budget comparison of mode orderings on Pythia-70M.}
Both orderings use the locked 51-checkpoint basis, two-node quadrature ($Q=2$), and the same 128 frozen Pile-test contexts.
Behavioral-energy ordering is ranked on the validation contexts, while parameter-energy ordering uses $\sigma_k^2$ and no analysis inputs.
The reported $R^2$ is the pooled centered-$\Delta$logit score in Equation~\eqref{eq:pythia-pooled-r2}, and KL is mean $D_{\mathrm{KL}}(p_{\mathrm{trained}}\Vert p_{\mathrm{recon}})$ in nats.}
\label{tab:pythia-ordering}
{\small
\begin{tabular}{@{}rrrrrrr@{}}
\toprule
& \multicolumn{2}{c}{pooled centered $\Delta$logit $R^2$} & \multicolumn{2}{c}{mean KL (nats)} & \multicolumn{2}{c}{Top-1 prediction agreement} \\
\cmidrule(lr){2-3}\cmidrule(lr){4-5}\cmidrule(l){6-7}
$K$ & \textbf{behavioral} & \textbf{parameter} & \textbf{behavioral} & \textbf{parameter} & \textbf{behavioral} & \textbf{parameter} \\
\midrule
$1$  & $0.451$  & $0.121$  & $76.6$  & $27.3$  & $0\%$    & $4.7\%$ \\
$3$  & $0.601$  & $0.133$  & $53.1$  & $21.4$  & $0\%$    & $6.3\%$ \\
$5$  & $0.728$  & $0.134$  & $58.3$  & $22.1$  & $0\%$    & $8.6\%$ \\
$10$ & $0.858$  & $0.741$  & $50.9$  & $41.7$  & $0\%$    & $0\%$ \\
$20$ & $0.9668$ & $0.9448$ & $27.4$  & $43.7$  & $0\%$    & $0\%$ \\
$50$ & $0.9998$ & $0.9998$ & $0.146$ & $0.146$ & $77.3\%$ & $77.3\%$ \\
\bottomrule
\end{tabular}}
\end{table}
\end{revblock}

\begin{figure}[h]
\centering
\includegraphics[width=0.92\linewidth]{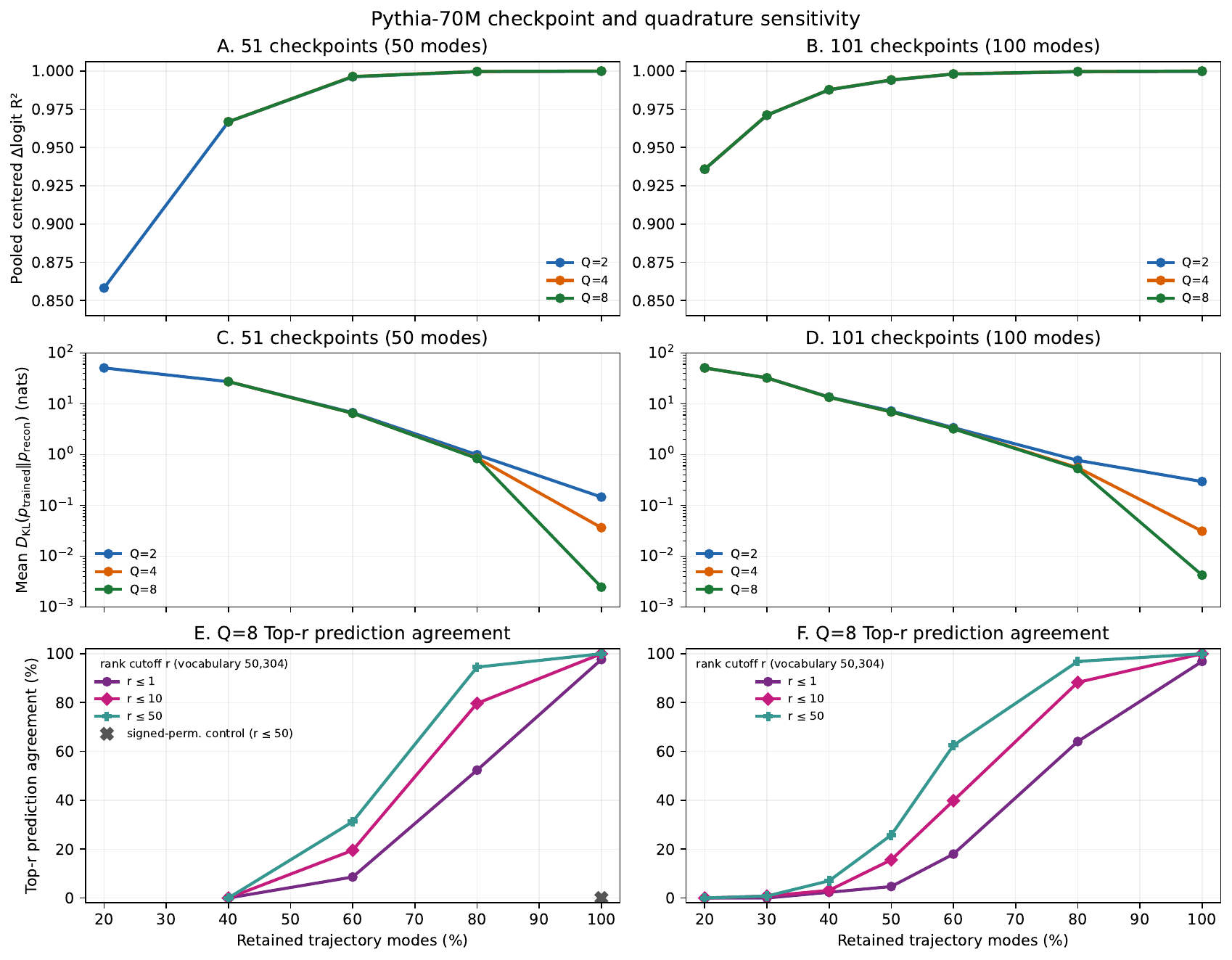}
\caption{\textbf{Post-hoc Pythia checkpoint-resolution and quadrature sensitivity.}
Columns hold the trajectory discretization fixed at 51 checkpoints (50 modes) or 101 nested checkpoints (100 modes).
Rows report pooled centered-$\Delta$logit $R^2$, mean full-vocabulary $D_{\mathrm{KL}}(p_{\mathrm{trained}}\Vert p_K)$ in nats on a log scale, and $Q=8$ Top-$r$ prediction agreement.
Lower KL is better.
Here $K$ is the retained-mode count, while Top-$r$ prediction agreement is top-$r$ accuracy with the trained model's highest-logit token in place of the ground-truth label: the fraction of contexts where that token lies among the reconstruction's $r\in\{1,10,50\}$ highest-logit candidates.
It is therefore neither true-token accuracy nor candidate-set intersection, and $r=1$ is exactly the Top-1 prediction agreement reported elsewhere.
Even $r=50$ covers only 0.10\% of the vocabulary.
Panel E also marks the signed-permutation control at full retention, scored at the same $Q=8$. It is defined only for the 51-checkpoint 50-mode path, so panel F has no counterpart.
The 51-checkpoint $Q=8$ sweep starts at $K=20$ modes (40\% retention), so panel E has no 20\% or 30\% points.
Within a schedule, quadrature leaves the mode order fixed.
The 101-checkpoint $Q=4,8$ endpoints use the exact checkpoint displacement, numerically equivalent to retaining all modes.}
\label{fig:supp:pythia-checkpoint-quadrature}
\end{figure}

\paragraph{Post-hoc checkpoint and quadrature sensitivity.}
These checks vary three analysis settings chosen after the locked result, namely checkpoint density, quadrature order, and retained mode budget, and leave the evaluation data untouched.
We formed a nested 101-checkpoint schedule that retains all 51 original checkpoints and adds 50 released snapshots, reducing every later interval to at most 2,000 optimizer steps.
This post-hoc schedule re-estimates the 100-mode basis and behavioral-energy ordering on the same validation ranking contexts, then scores the same frozen Pile-test contexts without test-set reranking.
Its flattened float32 checkpoint stack occupies approximately 28.5 GB, while path analysis scales with checkpoint intervals, retained directions, quadrature nodes, and scoring contexts.
An asymptotic checkpoint-density sweep was therefore computationally out of scope.
Table~\ref{tab:pythia-sensitivity} separates this checkpoint-resolution change from quadrature refinement on each fixed schedule, and Figure~\ref{fig:supp:pythia-checkpoint-quadrature} shows the corresponding retained-fraction curves.

\begin{table}[h]
\centering
\caption{\textbf{Post-hoc Pythia checkpoint and quadrature sensitivity.}
Concentration uses the fixed validation ranking contexts.
The $Q=2$ columns use the all-mode modal reconstruction, while $Q=4,8$ use the exact checkpoint displacement.
The all-mode endpoint differs from full-update only at numerical precision.
All output metrics use the same 128 frozen Pile-test contexts.
The reported $R^2$ is the pooled centered-$\Delta$logit score in Equation~\eqref{eq:pythia-pooled-r2}.
KL is mean $D_{\mathrm{KL}}(p_{\mathrm{trained}}\Vert p_{\mathrm{recon}})$ in nats.
Top-1 prediction agreement compares the trained model's and reconstruction's highest-logit tokens.
PR and $K_{\mathrm{eff}}$ fractions divide each statistic by the available mode count.}
\label{tab:pythia-sensitivity}
{\scriptsize
\setlength{\tabcolsep}{3.2pt}
\renewcommand{\arraystretch}{1.08}
\resizebox{\linewidth}{!}{%
\begin{tabular}{@{}lrrrrrr@{}}
\toprule
\textbf{Checkpoints} & \textbf{Modes} & \textbf{PR / modes} & \textbf{$K_{\rm eff}(0.9)$ / modes} & \textbf{\shortstack{Full-update pooled\\centered $\Delta$logit $R^2$\\$Q=2/4/8$}} & \textbf{\shortstack{Full-update mean KL (nats)\\$Q=2/4/8$}} & \textbf{\shortstack{Full-update Top-1\\prediction agreement\\$Q=2/4/8$}} \\
\midrule
51 (locked) & 50 & $5.09$ (10.2\%) & $13$ (26.0\%) & $0.999809/0.999966/0.999997$ & $0.1463/0.0366/0.00246$ & $77.3\%/95.3\%/97.7\%$ \\
101 (post-hoc nested) & 100 & $7.82$ (7.8\%) & $21$ (21.0\%) & $0.999705/0.999947/0.999994$ & $0.2954/0.0312/0.00425$ & $77.3\%/96.1\%/96.9\%$ \\
\bottomrule
\end{tabular}}}
\end{table}

The absolute PR and $K_{\mathrm{eff}}$ increase when twice as many update intervals make twice as many modes available, while their fractions remain of similar order and decrease modestly.
Table~\ref{tab:pythia-retained} compares the two schedules at matched retained fractions.
Thus individual modes and absolute counts are schedule-dependent, but both resolutions show the same qualitative progression from rapid centered-logit recovery to later decision recovery.

\begin{table}[h]
\centering
\caption{\textbf{Matched retained fractions across checkpoint schedules.}
Two-node ($Q=2$) held-out scores on the same 128 frozen Pile-test contexts, with modes ranked on the validation contexts.
Retained fraction is the kept mode count divided by the available mode count, so $80\%$ is 40 of 50 modes for the 51-checkpoint basis and 80 of 100 for the 101-checkpoint basis.
The reported $R^2$ is the pooled centered-$\Delta$logit score in Equation~\eqref{eq:pythia-pooled-r2}.}
\label{tab:pythia-retained}
{\small
\begin{tabular}{@{}lrrrr@{}}
\toprule
& \multicolumn{2}{c}{pooled centered $\Delta$logit $R^2$} & \multicolumn{2}{c}{Top-1 prediction agreement} \\
\cmidrule(lr){2-3}\cmidrule(l){4-5}
\textbf{Retained} & \textbf{51 ckpt} & \textbf{101 ckpt} & \textbf{51 ckpt} & \textbf{101 ckpt} \\
\midrule
$20\%$  & $0.8582$  & $0.9357$  & $0\%$    & $0\%$ \\
$40\%$  & $0.9668$  & $0.9877$  & $0\%$    & $1.6\%$ \\
$60\%$  & $0.9962$  & $0.99795$ & $10.9\%$ & $17.2\%$ \\
$80\%$  & $0.99958$ & $0.99947$ & $46.1\%$ & $59.4\%$ \\
$100\%$ & $0.99981$ & $0.99970$ & $77.3\%$ & $77.3\%$ \\
\bottomrule
\end{tabular}}
\end{table}
For full-update, the large improvement from $Q=2$ to $Q=4$ and the much smaller change from $Q=4$ to $Q=8$ identify quadrature as the dominant numerical error and show convergence on both checkpoint schedules.
Across the dense truncated set $K\in\{20,30,40,50,60,80\}$, the $Q=4$ and $Q=8$ curves differ by at most $3.6\times10^{-5}$ in pooled centered-$\Delta$logit $R^2$ and one of 128 contexts in Top-1 prediction agreement.
At $K=80$, the dense $Q=2/4/8$ results are pooled centered-$\Delta$logit $R^2=0.999472/0.999645/0.999681$ and Top-1 prediction agreement $59.4\%/64.1\%/64.1\%$.
The corresponding mean KL values are $0.773/0.554/0.531$ nats.
At $Q=8$, perplexity is $56.00$ versus $35.04$ for the trained model, while full-update KL and perplexity are $0.00425$ nats and $34.94$.
Thus the remaining fixed-schedule decision gap at this retained fraction is truncation-sensitive rather than a failure of quadrature convergence.

\begin{figure}[h]
\centering
\includegraphics[width=0.92\linewidth]{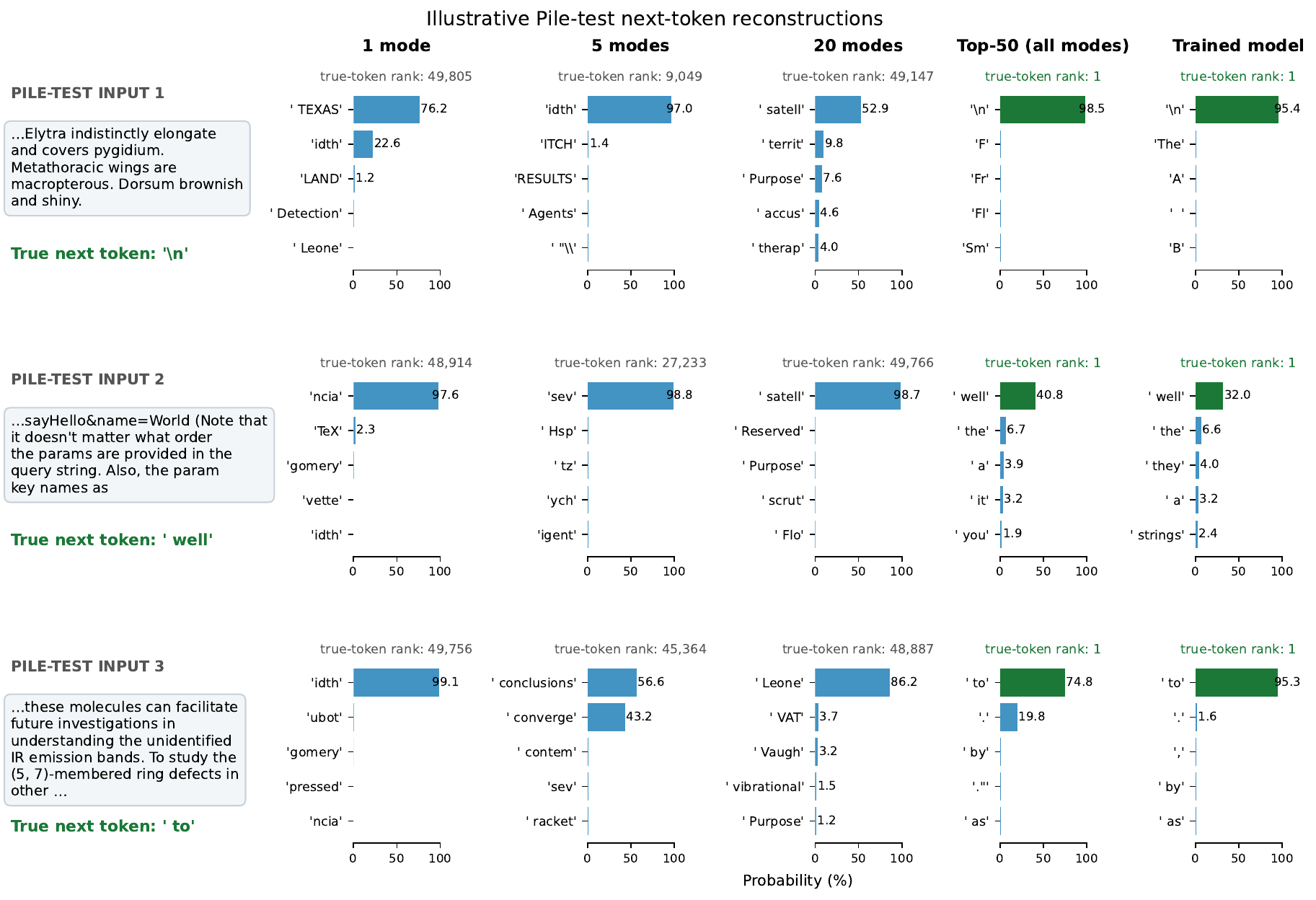}
\caption{\textbf{Illustrative untouched-test next-token distributions across cumulative modal reconstructions.}
Each row uses one Pile-test scoring context.
Columns show cumulative 1-, 5-, 20-, and 50-mode reconstructions followed by the trained-model reference.
Bars show the five highest-probability tokens, green marks the true next token, and each panel reports its rank.
Examples were selected deterministically among contexts the trained model predicts correctly.
The reported true-token rank is therefore also the rank of the trained model's predicted token and measures the same quantity that Figure~\ref{fig:supp:pythia-checkpoint-quadrature}E--F aggregates as Top-$r$ prediction agreement, here for the locked two-node analysis rather than the eight-node sweep.
These examples are explanatory rather than aggregate evidence.}
\label{fig:supp:pythia-examples}
\end{figure}

Figure~\ref{fig:supp:pythia-examples} illustrates why high pooled centered-$\Delta$logit $R^2$ can coexist with decision errors.
Under the locked two-point rule, the 20-mode reconstruction has zero Top-1 prediction agreement across the 128 scoring contexts, while the 50-mode along-path reconstruction reaches $77.3\%$ and has perplexity $38.66$.
A near-perfect pooled centered-$\Delta$logit $R^2$ therefore cannot certify decision recovery, because small errors near competing-token boundaries can change the highest-logit prediction.
Figure~\ref{fig:supp:pythia-checkpoint-quadrature}E--F relaxes the rank cutoff, and Table~\ref{tab:pythia-topr} gives the corresponding values at 80\% retention.
Both full-update endpoints reach 100\% by $r=3$.

\begin{table}[h]
\centering
\caption{\textbf{Top-$r$ prediction agreement at 80\% retention.}
Eight-node ($Q=8$) scores on the same 128 frozen Pile-test contexts, corresponding to Figure~\ref{fig:supp:pythia-checkpoint-quadrature}E--F.
A context counts as agreeing when the trained model's highest-logit token lies among the reconstruction's $r$ highest-logit candidates.}
\label{tab:pythia-topr}
{\small
\begin{tabular}{@{}lrrrrrr@{}}
\toprule
\textbf{Rank cutoff $r$} & $1$ & $3$ & $5$ & $10$ & $20$ & $50$ \\
\midrule
51 checkpoints, 40/50 modes   & $52.3\%$ & $67.2\%$ & $71.9\%$ & $79.7\%$ & $89.1\%$ & $94.5\%$ \\
101 checkpoints, 80/100 modes & $64.1\%$ & $77.3\%$ & $84.4\%$ & $88.3\%$ & $91.4\%$ & $96.9\%$ \\
\bottomrule
\end{tabular}}
\end{table}
Neither the signed-permutation control nor the 20-mode reconstruction ever places the trained model's token within the top 50 candidates on any of the 128 contexts, at either $Q=2$ or $Q=8$, whereas the locked 50-mode reconstruction reaches $77.3/94.5/100\%$ at $r=1/3/10$.
Eight-node quadrature moves the control only from $-0.890$ to $-0.891$ in pooled centered $\Delta$logit $R^2$, confirming that scrambling parameter coordinates fails structurally rather than through integration error.
This experiment supports compact reconstruction of centered logit-change variance for one public 70M-parameter trajectory.
It does not establish decision-preserving model compression, mode monosemanticity, causal necessity, or scaling behavior.

\end{document}